\documentclass[runningheads]{llncs}
\usepackage{standalone}
\usepackage{graphicx}
\usepackage{amsmath,amssymb}
\usepackage{hyperref}
\usepackage[font=scriptsize]{subfig}
\usepackage{tabularx}
\usepackage{amsfonts}
\usepackage{multirow}
\newcommand{\T}[2]{\mathbf{T}^{#1}_{#2}}
\newcommand{\Pl}[1]{\mathbf{\pi}_{#1}}
\newcommand{\x}[1]{\mathbf{x}_{#1}}
\newcommand{\Q}[1]{\mathbf{Q}^{*}_{#1}}
\newcommand{\C}[1]{\mathbf{C}^{*}_{#1}}
\usepackage{flushend}

\begin{document}

\title{Structure Aware SLAM using Quadrics and Planes\thanks{Supported by the ARC Laureate Fellowship FL130100102 to IR and the ACRV CE140100016.}}
\titlerunning{Structure Aware SLAM} 


\author{Mehdi Hosseinzadeh\inst{1,3} \and
Yasir Latif\inst{1,3} \and
Trung Pham\inst{4} \and
Niko Suenderhauf\inst{2,3} \and
Ian Reid\inst{1,3}}
%

\authorrunning{M. Hosseinzadeh et al.} 


\institute{The University of Adelaide, Adelaide, Australia 
\email{\{firstname.lastname\}@adelaide.edu.au}
\and 
Queensland University of Technology, Brisbane, Australia
\email{niko.suenderhauf@qut.edu.au}\\
\and
Australian Centre for Robotic Vision, Brisbane, Australia
\and
NVIDIA, Santa Clara CA 95051, USA \\
\email{trungp@nvidia.com}}

\maketitle

\begin{abstract}
Simultaneous Localization And Mapping (SLAM) is a fundamental problem in mobile robotics. 
While point-based SLAM methods provide accurate camera localization, the generated maps lack semantic information. On the other hand, state of the art object detection methods provide rich information about entities present in the scene from a single image.
This work marries the two and proposes a method for representing generic objects as quadrics which allows object detections to be seamlessly integrated in a SLAM framework. 
For scene coverage, additional dominant planar structures are modeled as infinite planes.
Experiments show that the proposed points-planes-quadrics representation can easily incorporate Manhattan and object affordance constraints, greatly improving camera localization and leading to semantically meaningful maps.
The performance of our SLAM system is demonstrated in \href{https://youtu.be/dR-rB9keF8M}{https://youtu.be/dR-rB9keF8M}.

\end{abstract}

\section{Introduction}\label{sec:intro}

Simultaneous Localization And Mapping (SLAM) is one of the fundamental problems in
mobile robotics \cite{cadena2016past}
and addresses the reconstruction of a previously
unseen environment while simultaneously localizing a mobile robot with respect to it. 
While the representation of the robot pose depends on the degrees of freedom of motion,
the representation of the map depends on a multitude of factors including 
the available sensors,
computational resources,
intended high level task,
and required precision. 
Many possible representations have been proposed. 

For visual-SLAM,
the simplest representation of the map is a collection of 3D points
which correspond to salient image feature points. 
This representation is sparse and efficient to compute and update.
Point based methods have been successfully used to map city-scale environments.
However, this sparsity comes at a price: points-based maps lack semantic information
and are not useful for high level task such as grasping and manipulation. 
Although methods to compute denser representations have been proposed  \cite{dso,lsdslam,dtam,infinitam,kinectfusion} these representations remain equivalent to a collection of points and therefore carry no additional semantic information.

Man-made environments contain many objects that could potentially be used as landmarks in a SLAM map, encapsulating a higher level of information than a set of points. Previous object-based SLAM efforts have mostly relied on a database of predefined objects -- which must be recognised and a precise 3D model fit to match the observation in the image to establish correspondence \cite{DBLP:conf/cvpr/Salas-MorenoNSKD13}. Other work \cite{Bao_CVPR2011_SSFM} has admitted more general objects (and constraints) but only in a slow, offline structure-from-motion context. In contrast, we are concerned with live (real-time) SLAM, but we seek to represent a wide variety of objects. Like \cite{Bao_CVPR2011_SSFM} we are not so concerned with high-fidelity reconstruction of individual objects, but rather to represent the location, orientation and rough shape of objects.  A suitable representation is therefore potentially a quadric \cite{sfmquadric,dualquadNiko2017arXiv}, which allows a compact representation of rough extent.

In addition to objects, much of the large-scale structure of a general scene (especially indoors) comprises dominant planar surfaces. Including planes in a SLAM map has also been explored before \cite{Salas-Moreno:ISMAR,kaess-plane}. Planes are also a good representation for feature deprived regions, where they provide information complimentary to points and can represent significant portions of the environment with very few parameters, leading to a representation that can be constructed and updated online \cite{kaess-plane}. Pertinent to our purpose, such a representation also provides the potential for additional constraints for the points that lie on one of the planes and permits the introduction of useful affordance constraints between objects and their supporting planes, as we explain later in the paper. All these constraints lead to better estimate of the camera pose.

Modern SLAM is usually formulated as an unconstrained sparse nonlinear least-square problem \cite{grisetti2010tutorial}. 
The sparsity structure of the problem greatly effects the computation time of the systems.
If planes and quadrics are to be introduced in a SLAM system, they should be represented in a way which is amenable to the non-linear least squares formulation
and respects the sparsity pattern of the SLAM problem.

In this work, we propose a map representation that consists of points, and higher level geometric entities such as planes and objects as landmarks.  Unlike previous work such as 
\cite{Bao_CVPR2011_SSFM} we explicitly target real-time performance, and integrate within an online SLAM framework. Such performance would be impossible with uncritical choices of representation and to that end we propose a novel representation of objects based on quadrics that decomposes to permit clean, fast and effective real-time implementation. We show that this representation, along with point-plane, plane-plane (Manhattan), and plane-object (supporting) constraints, greatly reduces the error in the estimated camera trajectory without incurring great extra computational cost. Because of the higher-level primitives in the map, the representation remains compact, but carries crucial semantic information about the scene.
To the best of authors' knowledge, this is the first \textit{real-time} SLAM system proposed in literature that incorporates both higher level primitives of planes and previously unseen objects as landmarks.

The main contributions of the paper are articulated as follows: (1) proposing a novel representation and decomposition of dual quadrics, and its related factors for integrating objects in the SLAM factor graph that is amenable to the non-linear least-squares framework and allows CNN-based object detections to be seamlessly integrated in a SLAM framework, (2) introducing a supporting affordance relationship between quadric objects and planes in a SLAM factor graph thanks to the proposed representation, and (3) integrating all of the higher level primitives: planes and quadrics, along with points, and geometric relationships among them, Manhattan assumptions and supporting/tangency constraints, in a complete online key-frame based SLAM system that performs near real-time. 

The remainder of the paper is organized as follows.
In the next section, we present the background for SLAM as the solution to a factor-graph, and explain how our proposal is integrated into such a framework. In particular, we give detailed descriptions of the mathematical representations of each landmark, and the factors they induce. Section \ref{sec:slam_system} presents an overview of how the preceding section is integrated into an overall SLAM system. Experiments showing the efficacy and comparative performance of our system are presented in Section
\ref{sec:experiments}. We conclude with a summary and discussion of future research directions.

\section{Related Work}\label{sec:relatedwork}

SLAM is well studied problem in mobile robotics and many different solutions have been proposed for solving it. 
The most recent of these is the graph-based approach that formulates SLAM as a nonlinear least squares problem 
\cite{grisetti2010tutorial}. 
SLAM with cameras has also seen advancement in theory and good implementations that have lead to many real-time
systems from sparse (\cite{orbslam},\cite{dso}) to semi-dense (\cite{lsdslam}, \cite{svo}) to fully dense
(\cite{dtam}, \cite{kinectfusion}, \cite{infinitam}).

Recently, there has been a lot of interest in extending the capability of a point-based representation by either applying the same techniques to other geometric primitives or fusing points with lines or planes to get better accuracy. Several methods have explored replacing points with lines 
\cite{lemaire2007monocular,gee2006real}.
However, lines present especial difficulty because of the lack of a good mathematical representation that is amenable to the least-squares framework. 
Some works have explored the possibility of using lines and points in the same framework \cite{pumarola2017pl,plslam2017} and have been more successful. 

Recently, \cite{kaess-plane} proposed a representation for infinite planes that is amenable for use in a least-squares framework. Using this representation, they presented a method that work using just information of planes visible in the environment. Similarly, \cite{yang2016pop} use a monocular input to generate plane hypothesis using a Convolutional Neural Network (CNN) which is then refined over time using both the planes as well as points in the images. \cite{taguchi2013point} proposed a method that fuses points and planes using an RGB-D sensor. In the latter works, they try to fuse the information of plane entities to increase the accuracy of depth inference.

Quadrics based representation was first proposed in \cite{cross1998quadric} and later used in a structure from motion setup \cite{sfmquadric}. \cite{sunderhauf2017meaningful} presented a semantic mapping system that uses object detection coupled with RGB-D SLAM to reconstruct precise object models in the environment, however object models do not inform localization. \cite{DBLP:conf/cvpr/Salas-MorenoNSKD13} presented an object based SLAM system that uses pre-scanned object models as landmarks for SLAM but can not be generalized to unseen objects.
\cite{mccormac2017semanticfusion} presented a system that fused multiple semantic predictions with a dense map reconstruction. SLAM is used as the backbone to establish multiple-view correspondences for fusion of semantic labels without informing the localization.

\section{Landmark Representations}\label{sec:representations}

For object-oriented SLAM the map comprises not only points but higher-level entities representing landmarks which aim to be more semantically meaningful than sparse points. However to maintain real-time operation, there is a trade-off between complexity of the landmark representation and the computational cost of tracking and mapping. 

In this work we consider two kinds of landmarks, which admit efficient implementation but can broadly capture the overall structure of many scenes, especially those captured indoors:  \textbf{a)} plane landmarks, whose role is to encapsulate high-level structure of regions; and \textbf{b)} quadrics (more specifically ellipsoids) that serve as a general representation of objects in scene, capturing not detailed shape, but key properties such as size, extent, position and orientation. We introduce representations for both types of primitive that allow for efficient implementation in a SLAM framework, as well as admitting clean and effective constraints between primitives, such as supporting constraint between objects and planes.

\subsection{Quadric Representation}\label{subsec:quadric}
As noted above, we represent general objects in a scene using an ellipsoid. Generally speaking, a quadric surface in 3D space can be represented by a homogeneous quadratic form defined on the 3D projective space $ \mathbb{P}^{3} $ which satisfies $ \mathbf{x^{T}Qx=0} $, where $ \mathbf{x} \in \mathbb{R}^{4} $ is the homogeneous 3D point and $ \mathbf{Q} \in \mathbb{R}^{4\times4} $ is the symmetric matrix representing the quadric surface. However the  
relationship between a point-quadric and its projection into a camera (a conic) is not straightforward \cite{Hartley:2003:MVG:861369}. A widely accepted alternative is to make use of the dual space  \cite{cross1998quadric,sfmquadric,dualquadNiko2017arXiv} in which a quadric is represented as the envelope of a set of tangent planes, viz: 
\begin{equation}
\Pl{}^{T}\Q{}\Pl{}=0
\end{equation}

This greatly simplifies the relationship between the quadric and its projection to a conic, however a further problem remains in the context of optimisation in a graph-SLAM framework. The issue is that an update of $ \Q{} $, given an 9-dim error vector $ \mathbf{e} $ in the tangent space of $ \Q{} $, should be constrained to lie along a geodesic of the manifold. But finding these geodesics and updating with respect to them is computationally expensive, making a ``straightforward'' quadric representation intractable for incremental optimisation. 

We seek to address both of these issues. For our object representation, we would like to restrict landmarks to belong to the set of bounded quadrics, namely ellipsoids. To do so requires imposing the constraint that $\Q{}$ must have 3 positive and 1 negative eigenvalues.
Based on this restriction, the representation of dual ellipsoids $ \Q{} $ can be decomposed as: 
\begin{equation}
\Q{} = \T{}{Q} \Q{c} \T{T}{Q} = 
\begin{bmatrix} \mathbf{R} & \mathbf{t} \\ \mathbf{0}^{T} & 1 \end{bmatrix}
\begin{bmatrix} a^{2} & 0 & 0 & 0 \\ 
				0 & b^{2} & 0 & 0 \\
				0 & 0 & c^{2} & 0 \\
				0 & 0 & 0 & -1 	  \\
\end{bmatrix}
\begin{bmatrix} \mathbf{R}^{T} & \mathbf{0} \\ \mathbf{t}^{T} & 1 \end{bmatrix}
\label{eq:quadric_decompose}
\end{equation}
\noindent
where $ \T{}{Q} \in \mathbf{SE(3)}$ transforms an axis-aligned (canonical) quadric at the origin, $ \Q{c} $, to a desired $ \mathbf{SE(3)} $ pose, and $ a $, $ b $, $ c $ denote the scale of the canonical ellipsoid $ \Q{c} $ along its principal axes.

Optimizing on the space of quadrics must impose constraints on the eignevalues of $ \Q{} $ to force the solution to be an ellipsoid. Recently \cite{quadobject} and \cite{sfmquadric} have parameterized ellipsoids to overcome this problem. They optimize on the space of ellipsoids, $ \mathbb{E} $, to localise the quadric by their respective conic observations. However their representation requires solving a constrained least squares problem.
While their parametrization is useful for observations of quadrics on camera frames as conics, it can not be used as generic constraints in the graph SLAM problem due to its constrained nature. 
The authors in \cite{sfmquadric} decompose the translation part of the representation, mainly for numerical stability in the optimisation because of the different scales of translation and the other parts of the representation, and impose some prior knowledge on the shape of the ellipsoids. 
For a more efficient representation of ellipsoids in graph-based SLAM, we exploit the underlying structure of $ \mathbb{E}$ to represent the dual quadric as follows:
\begin{equation}
\Q{} = \T{}{Q} ~ \Q{c} ~ \T{T}{Q} = 
\begin{bmatrix} \mathbf{R} & \mathbf{t} \\ \mathbf{0}^{T} & 1 \end{bmatrix}
\begin{bmatrix} \mathbf{LL}^{T} & \mathbf{0} \\ \mathbf{0}  & -1 \end{bmatrix}
\begin{bmatrix} \mathbf{R}^{T} & \mathbf{0} \\ \mathbf{t}^{T} & 1 \end{bmatrix}
\quad\text{where}\quad
\mathbf{L} = 
\begin{bmatrix} a & 0 & 0 \\ 
				0 & b & 0 \\
				0 & 0 & c
\end{bmatrix}				
\end{equation}
\noindent
with real numbers $a$, $b$ and $c$, and so $ \mathbf{LL}^{T} $ guarantees the required positive eigenvalues. 
We thus represent a dual ellipsoid $ \Q{} $ using a tuple $ \mathbf{(T,L)} $ where $ \mathbf{T \in SE(3)} $ 
and $ \mathbf{L} $ lives in $ \textbf{D(3)} $ the space of real diagonal $ 3 \times 3 $ matrices, i.e. an axis-aligned ellipsoid accompanied by a rigid transformation. This decomposition exploits the underlying $ \mathbf{SE(3) \times D(3)} $ structure of the manifold of $ \mathbb{E} $, ensuring we remain in the space of ellipsoids without needing to solve a constrained optimisation problem. 

We update the $ \Q{} = \mathbf{(T,L)} $ separately in the underlying 6D space of $ \mathbf{SE(3)} $ and 3D space of $ \mathbf{D(3)} $, where both of them are Lie groups and can be updated efficiently by their respective Lie algebra. Thus the proposed update rule is: 
\begin{equation}
\mathbf{\Q{} \oplus \varDelta\Q{} = (T,L) \oplus (\varDelta T, \varDelta L) = (T \cdot \varDelta T, L + \varDelta L)}
\end{equation}
where $ \mathbf{\oplus:\mathbb{E}\times\ \mathbb{E} \longmapsto \mathbb{E}} $ is the mapping for updating ellipsoids, $ \mathbf{\varDelta L} $ is the update for $ \mathbf{L} $ which comes from the first 3 elements of error vector $ \mathbf{e} $ and applies in the Euclidean space of $ \mathbb{R}^{3} $ and $ \mathbf{\varDelta T} $  is the update for $ \mathbf{T} $ which comes from the last 6 elements of error vector $ \mathbf{e} $ and applies in space of $ \mathbf{SE(3)} $.
This decoupled update is a good approximation given the incremental nature of evidence. 

This proposed representation of ellipsoids is beneficial particularly when we want to impose constraints on different parts of this representation. 
For instance, this representation for $ \Q{} $ makes it possible to apply prior knowledge for shapes and sizes of objects, using the $\mathbf{L}$ component, prior information about location and orientation of the object using the $\mathbf{T}$ component, and adjacency/supporting constraints (see Section \ref{subsec:constraints}).

\subsection{Plane Representation}\label{subsec:plane}
To represent planes as structural entities in the map, we represent an  infinite plane $ \Pl{} $ by its normalised homogeneous coordinates $ \Pl{} = \left[a~b~c~d\right]^{T} $ where $ \vec{\mathbf{n}} = \left[a~b~c\right]^{T} $ is the normal vector and $d$ is the distance to origin. The reason for considering normalised homogeneous vectors is inspired by \cite{kaess-plane} to have a minimal representation for planes to avoid rank-deficient information matrices in optimization. 
This representation of the planes is isomorphic to the northern hemisphere of $ \mathcal{S}^{3} $, 
or equivalently the $\mathbf{SO}(3)$ Lie group, therefore the optimisation can be performed using three elements that represent an element of $\mathbf{SO}(3)$.

\subsection{SLAM as a factor-graph} \label{subsec:constraints}
Following the seminal work of \cite{Dellaert} it is now well known that SLAM can be represented as a factor graph $\mathcal{G}(\mathcal{V},\mathcal{E})$
where the vertices $\mathcal{V}$ represent the variables that need to be estimated such as robot poses and points in 3D,
and the edges $\mathcal{E}$ represent constraints or \textit{factors} between the vertices.

In a traditional point-based SLAM system, factors exist between points and the camera that seek to minimize reprojection error:
\begin{equation}
f_r(\mathbf{x}_w, \mathbf{T}_{c}^{w}) = \parallel \mathbf{u}_c - \Pi(\mathbf{x}_w, \mathbf{T}_c^{w}) \parallel_{\mathbf{\Sigma}_r}
\end{equation}
\noindent
where $\mathbf{x}_w$ represent a point in the world,
$\mathbf{T}_{c}^{w}$ is the pose of the camera which takes a point in the current camera frame ($\mathbf{P}_c$) to a point in the world frame $\mathbf{P}_w = {\mathbf{T}_{c}^{w}}{\mathbf{P}_c}$ that is observed at the pixel location $\mathbf{u}_c$, and
$\Pi(.)$ is a function that projects a world point into the camera. $\parallel\mathbf{x}\parallel_{\mathbf{\Sigma}} $ is the mahalanobis norm and equal to $ \mathbf{x}^T\mathbf{\Sigma}^{-1}\mathbf{x}$ where $\mathbf{\Sigma}$ is the covariance matrix associated with the factor.
Likewise if odometry is known between two robot positions, a factor involving robot poses can be formulated as:
\begin{equation}
f_o( \mathbf{T}_{c}^{w}, \mathbf{T}_{k}^{w}) = 
\parallel  \mathbf{T}_{c,odom}^{k} \ominus \mathbf{T}_{c}^{k} \parallel_{\mathbf{\Sigma}_o}
\end{equation}
The solution to the SLAM problem is a configuration of the vertices $\mathcal{V}^{*}$ that minimizes the error over all the involved factors.

\begin{figure}[t]
\centering
\subfloat{\includegraphics[width=0.42\textwidth]{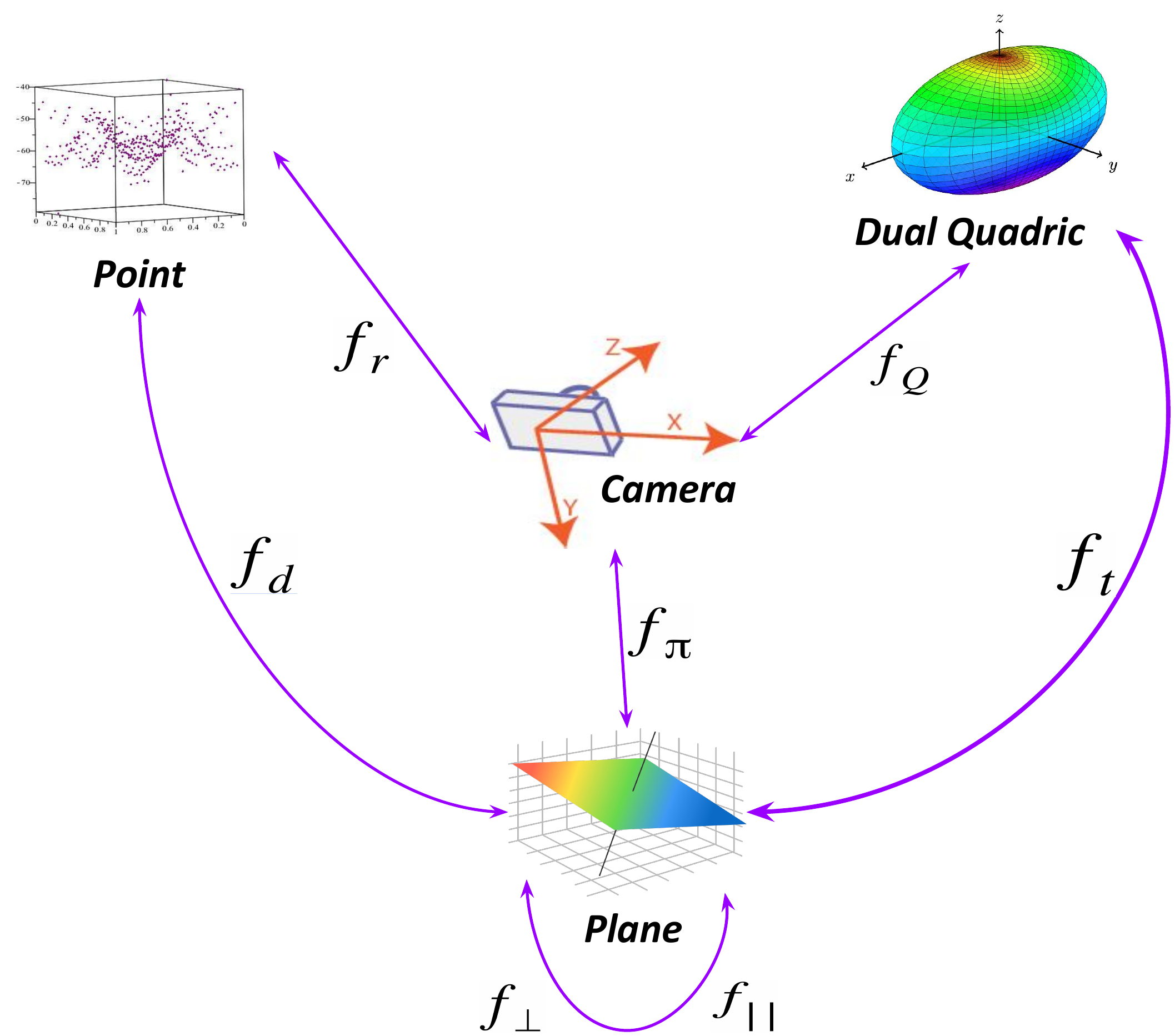}}
\caption{\small{The factor graph of our object-oriented SLAM system demonstrating all types of our landmark representations as nodes and observations and constraints as factors. For further details regarding these factors refer to section \ref{subsec:constraints}}}
\label{fig:facgraph}
\end{figure}

In our proposed object-oriented SLAM representation, the vertices in the SLAM graph consists not only of points but potentially planes and/or general objects (represented by quadrics). Fig.~\ref{fig:facgraph} shows the various factors involving cameras, points, planes, and quadric objects in our system. Below we describe in more detail how the new components of our SLAM system are introduced as additional factors in the graph. 

\subsubsection{Observations of Objects (Ellipsoids).}
A quadric $ \Q{} $ in the scene projects to a conic $ \C{} $ in an image \cite{Hartley:2003:MVG:861369}:
\begin{equation}
\begin{split}
\C{} \sim \mathbf{P \Q{} P^{T}} 
\quad \text{and} \quad
\mathbf{ P = K \begin{bmatrix} I_{3\times3} & 0_{3\times3} \end{bmatrix} \T{w}{c} } 
\end{split}
\end{equation}
where $ \mathbf{P} $ is the projection matrix of the camera with calibration matrix $ \mathbf{K} $ and $ \mathbf{\T{w}{c}} $ is the pose of the camera.
For observed conic $ \C{obs} $, we consider the observation error for quadric $ \Q{} $ as the Frobenius norm of the difference between normalized $ \C{obs} $ and normalized projected conic $ \C{} $:
\begin{equation}
f_Q(\Q{}, \mathbf{\T{w}{c}})=
\mathbf{ \parallel \C{} - \C{obs} \parallel_{F} = \sqrt{Tr((\C{} - \C{obs})(\C{} - \C{obs})^{T})} }
\end{equation}
\noindent
which forms a factor between the quadric and the camera pose.

\subsubsection{Observations of Planes.}
If we denote the observation of the plane $ \Pl{} $ from a camera pose $ \T{w}{c} $ by $ \Pl{obs} $, we can measure the observation error by:
\begin{equation}
f_{\pi}(\pi, \mathbf{\T{w}{c}}) = {{\parallel d({\T{w}{c}}^{-T}{\Pl{}} , \Pl{obs}) \parallel}_{\Sigma}^{2}}
\end{equation}
where $ {\T{w}{c}}^{-T}{\Pl{}} $ is the transformed plane to the camera coordinates frame and $ d $ is the distance function in the tangent space of the $\mathbf{SO}(3)$. For more details regarding plane updates and their corresponding exponential map refer to \cite{kaess-plane}. 

\subsubsection{Point-Plane Constraints.}
If we believe that a point actually lies on a specific plane, it makes sense to impose a constraint between the point and the relevant plane landmark. To do so we introduce the following factor:
\begin{equation}
f_{d}(x, \pi)={{\parallel \vec{\mathbf{n}}^{T}(\mathbf{x}-\x{o}) \parallel}_{\sigma}^{2}}
\end{equation}
which simply measures the orthogonal distance of the point $ \mathbf{x} $ from the plane $\pi$ with the unit normal vector $ \vec{\mathbf{n}} $. $ \x{o} $ is an arbitrary point on the infinite plane. 

\subsubsection{Plane-Plane Constraints (Manhattan Assumption).}
Imposing constraints on relative plane orientations is simply a matter of introducing a factor on the plane surface normals. The most useful and common such constraints (especially indoors) are those associated with a Manhattan world in which planes are mostly mutually orthogonal or parallel. Constraints between planes $ \Pl{1} $ and $ \Pl{2} $ with unit normal vectors $ \vec{\mathbf{n}_{1}} $ and $ \vec{\mathbf{n}_{2}} $, respectively, are implemented as:
\begin{align}
f_{\parallel}(\Pl{1}, \Pl{2}) & = {\parallel | \vec{\mathbf{n}}_{1}^\top \vec{\mathbf{n}}_{2} | - 1 \parallel}_{\sigma}^{2}   \qquad \text{\it for~parallel~planes} \quad \\
f_{\perp}(\Pl{1}, \Pl{2}) & = {\parallel \vec{\mathbf{n}}_{1}^\top \vec{\mathbf{n}}_{2} \parallel}_{\sigma}^{2} \qquad \text{\it for~perpendicular~planes} \quad
\end{align}

\subsubsection{Supporting/Tangency Constraints.}
Almost all stable objects in the scene are supported by structural entities of the scene like planes; e.g. commonly objects are found on the floor or on a desk. 
Such an affordance relationship can be imposed between a quadric object and a structural infinite plane by introducing a geometric tangency constraint between them. To the best of our knowledge, this is the first time  that such a constraint has been included in an online SLAM. 

Although imposing a tangency constraint in the space of point quadrics could be tricky, in the dual space such a constraint takes a particularly simple form:
\begin{equation}
f_t(\Pl, \Q{})={\parallel \Pl{}^{T}\Q{}\Pl{} \parallel}_{\sigma}^{2}
\end{equation}
where $ \Pl{} $ is the normalised homogeneous plane that supports the quadric $ \Q{} $.

\section{System Implementation}\label{sec:slam_system}
Modern SLAM system can be divided into two functional parts:
\textbf{a)} a front-end: which deals with raw sensory input to initialize vertices and factors and 
\textbf{b)} a back-end which optimizes the SLAM graph to create an updated estimate of the vertices.
In this section, we provide an overview of our front-end that extracts the landmarks, observations and constraints mentioned in the section \ref{sec:representations} to construct the SLAM graph.
\begin{figure}[t]
\centering
\subfloat{\includegraphics[width=0.8\textwidth]{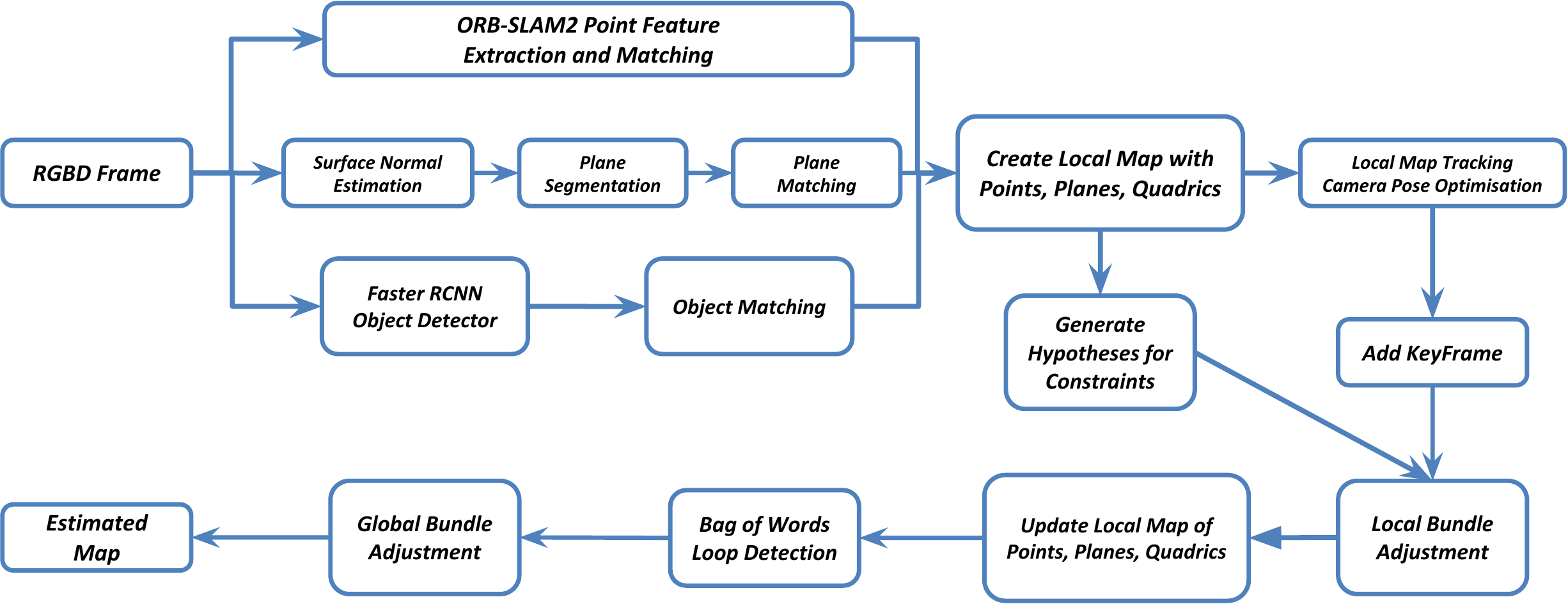}}
\caption{\small{The pipeline of our object-oriented SLAM system}}
\label{fig:system}
\end{figure} 
The back-end of our SLAM system, optimises this graph using a least-squares framework \cite{kummerle2011g}. 
It should be pointed out that
all of the landmarks and constraints participate in the optimisation after adding a new key-frame,
as well as when a loop closure is detected. 
Our system augments RGB-D variant of the publicly available ORB-SLAM2 \cite{orbslam}.
Loops are detected using bag of words \cite{galvez2012bags} based on ORB features. The pipeline of our system is demonstrated in Fig.~\ref{fig:system}.

\subsubsection{Point Observations.}
We rely on the underlying ORB-SLAM2 RGB-D implementation for points; candidate features are extracted based on uniqueness and described using ORB features, with their depth initialized using the depth channel of the input. For data-association across frames,
ORB features are matched in a coarse-to-fine pyramid in a local window around the previous observation.

\subsubsection{Plane Observations.}
For planar landmarks, we are interested not only in the parameters of the infinite planes, but also their extent visible in the current image, so that points can be associated to the planes on which they are observed.

Most plane fitting models for RGB-D data use RANSAC which is extremely slow for the purpose of building a near real-time online SLAM framework. 
Our plane segmentation follows \cite{trevor2013efficient} which segments point clouds from RGB-D data in near real-time.
For data-association across frames, we rely on the sparsity (few dominant planes in the scene) and inherent robustness (little variation frame-to-frame) of these landmarks. 
Using the difference between normals and the distance between planes,
data-association is done in a nearest-neighbor fashion.

The plane segmentation and matching uses depth data, and is the only part of our system (other than ORB feature depth initialization) which relies on depth information. In the future we aim to remove even this requirement and make the system truly monocular by hypothesizing planes using single-view semantic segmentation, depth and normal estimation, as is now possible by deep nets \cite{DBLP:conf/iccv/EigenF15}. 

\subsubsection{Conic Observations.}
We use Faster-RCNN~\cite{fasterrcnn} with pre-trained model on COCO dataset~\cite{coco} to detect the bounding boxes for objects of the scene. 
From the axis-aligned bounding boxes, the inscribed ellipsoid is computed as the conic projection of an observed quadric object. 
To avoid outliers and achieve robust detections we consider objects with 95\% or more detection confidence. 

For data-association across frames, we utilize the semantic labels and rely on the detection of the object to match the corresponding landmark. If more than one instance of a semantic class is found,  we use nearest-neighbor matching in the feature space generated by the detector. This simple strategy is successful with high-confidence object detections, as shown in section \ref{sec:experiments}.

Note that the partial occlusions can result in an inconsistent observations of a same object from different viewpoints that can lead to inaccuracy in the trajectory and map. The following course of actions is employed in our system to mitigate the negative impact of partial occlusions: \textbf{(a)} we use robust kernels (Huber) to robustify against large error, \textbf{(b)} only consider objects with 95\% or more detection confidence. With these two recourses we have seen almost consistent observations of COCO objects in our experiments shown in section \ref{sec:experiments}.

\subsubsection{Point-Plane Constraints.}
Finding association between points and planes is established during plane detection and segmentation. After detecting each plane and its finite boundary, its inlier points are determined to be those satisfying a threshold, $ th_{PP} $ distance, which we set as a function of the distance of the points from the camera, because further points have greater uncertainty. 

\subsubsection{Plane/Manhattan Constraints.}
The number of planes detected by our system is sufficiently small that we can consider all possible pairs, and introduce constraints with very little impact on overall speed of operation. At present we adopt the expedient of imposing a parallel constraint if the angle between the pair of planes is less than a threshold $th^{\parallel}_{M} $, and if the angle is within $th^{\bot}_{M}$ of 90 deg we introduce a perpendicular factor. For our experiments we have used $ th^{\parallel}_{M} = 15\deg $ and $ th^{\bot}_{M} = 75\deg $ in our system.

Manhattan constraints are imposed in a conservative manner with a large uncertainty and act as a prior on the relative orientation of the planes. Based on evidence gathered over image frames, they might end up being perpendicular or parallel but are not forced to be in that configuration if the data strongly favors an opposite interpretation.

\subsubsection{Supporting/Tangency Constraints.}
A supporting/tangency constraint between a quadric and a plane is imposed based on the orthogonal distance of the centroid of the quadric and infinite plane. If this distance is less than $ th_{S} $ we enforce the tangency constraint. In our experiments this threshold depends on the size of the quadric $ th_{S} = \max(20cm,a,b,c) $ where $ a $, $ b $, and $ c $ are half the length of the principal axes of the ellipsoid.

\section{Experiments}\label{sec:experiments}

We evaluate the performance of our SLAM system using the benchmarks RGB-D TUM dataset \cite{tum-dataset} and NYU-Depth V2 dataset \cite{nyu}. These sequences have a wide range of conditions,
from plane-rich scenes to scenes with little or no texture and also scenes with common objects such as those available in COCO dataset \cite{coco}.
We show qualitative as well as quantitative results of our system using different combinations of the proposed landmarks and constraints and compare the accuracy in the estimated camera trajectory against the RGB-D variant of the state-of-the-art sparse mapping system, ORB-SLAM2 \cite{orbslam}.

\subsection{Qualitative Results}

Some sequences in the TUM RGBD dataset contain little or no texture which makes it difficult for point-based SLAM systems to extract and track key-points. However these sequences have rich planar structures which are exploited by our SLAM system. 
The results for using planes with Manhattan constraints on \texttt{fr3/str\_notex\_far}
and \texttt{fr1/floor} are given in Fig.~\ref{fig:planes_experiments}. Results for more sequences are reported in the supplementary material.
The figure depicts the image frame along with tracked features, detected and segmented planes, and the reconstructed map consisting of points and planes from two different viewpoints. 
For these sequences, ORB-SLAM2 is unable to detect features in the environment with the normal
settings and loses track. Lowering the feature detection threshold in ORB-SLAM2 yields a greater number of features, but also results in more outliers leading to more inaccurate trajectories. 

\begin{figure}[t]
\centering
\subfloat{\includegraphics[width=0.24\textwidth]{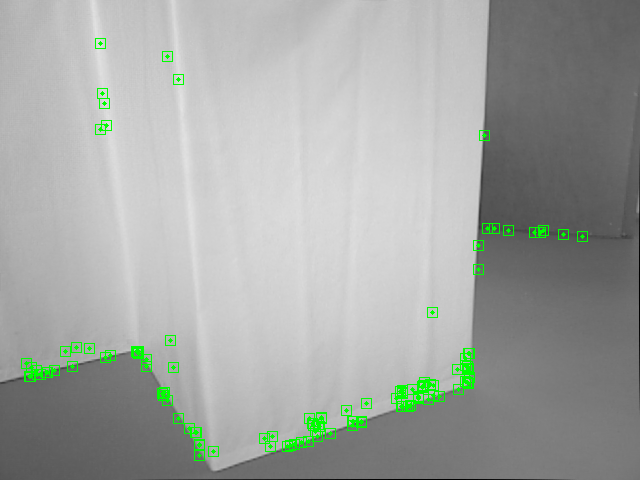}}~
\subfloat{\includegraphics[width=0.24\textwidth]{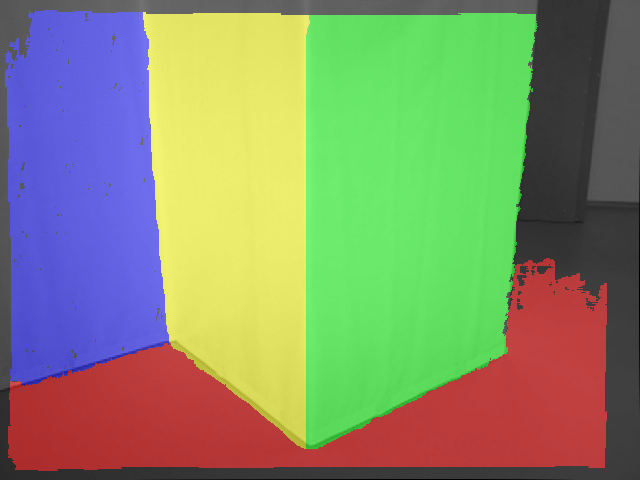}}~
\subfloat{\includegraphics[width=0.24\textwidth]{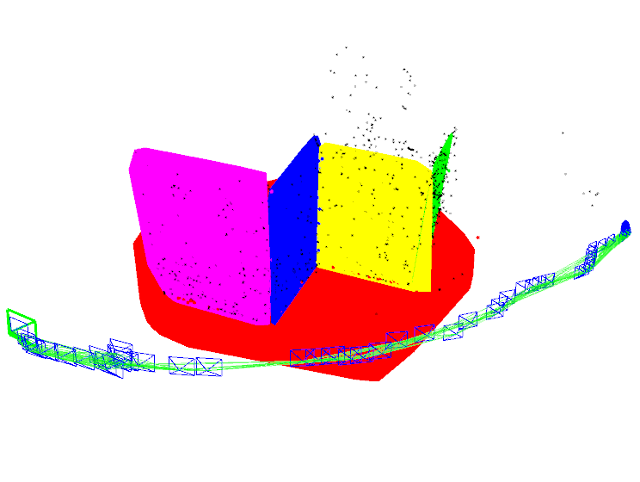}}~
\subfloat{\includegraphics[width=0.24\textwidth]{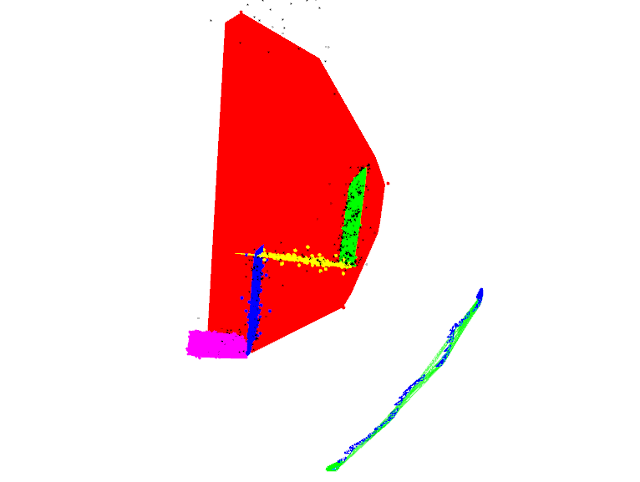}}\\[-2ex]
\setcounter{subfigure}{0}
\subfloat[ORB Features]{\includegraphics[width=0.24\textwidth]{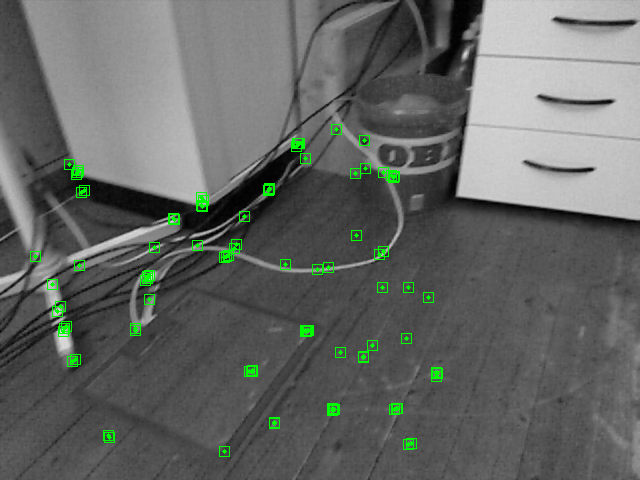}}~
\subfloat[Detected Planes]{\includegraphics[width=0.24\textwidth]{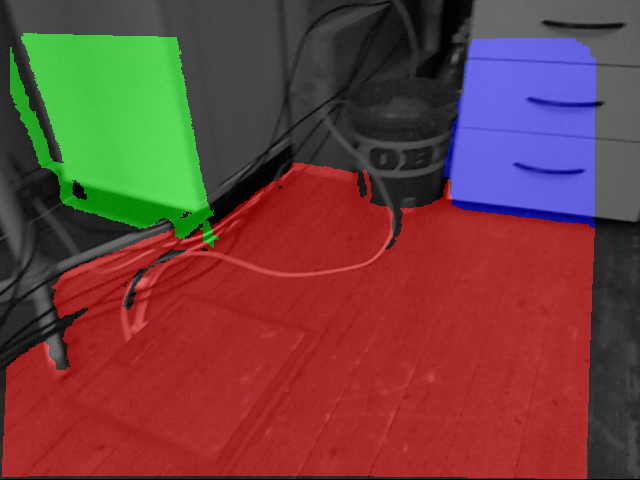}}~
\subfloat[Map (Side)]{\includegraphics[width=0.24\textwidth]{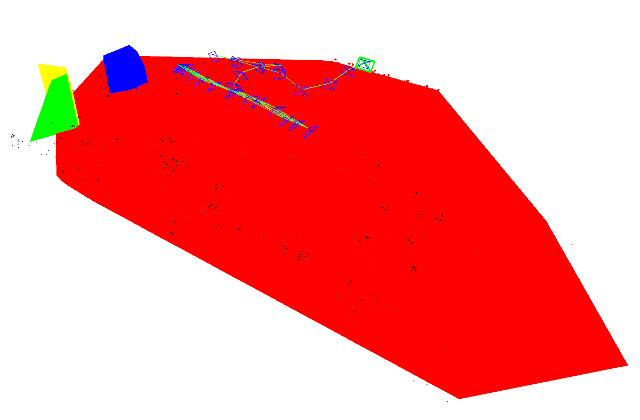}}~
\subfloat[Map (Top)]{\includegraphics[width=0.24\textwidth]{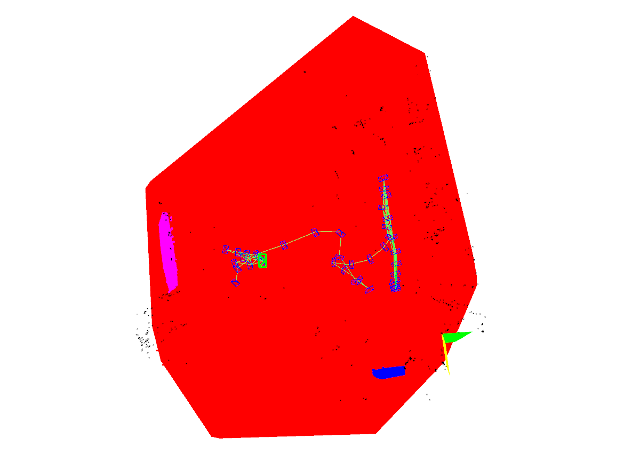}}
\caption{\small{Qualitative results for 2 different TUM RGBD datasets with low texture object-less but rich planar structures}}
\label{fig:planes_experiments} 
\end{figure}

\begin{figure}[t]
\centering
\subfloat{\includegraphics[width=0.24\textwidth]{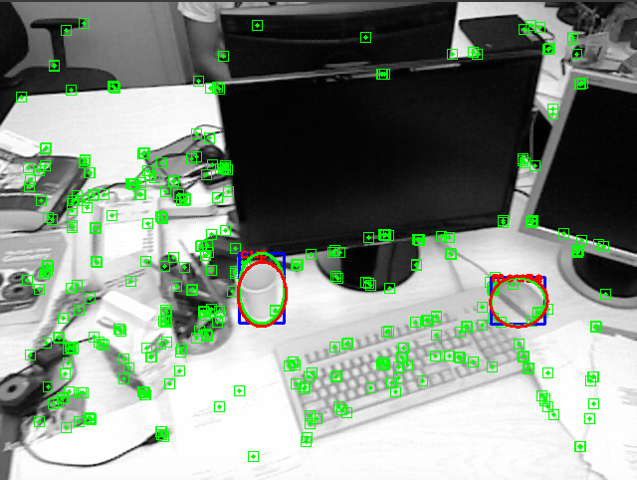}}~
\subfloat{\includegraphics[width=0.24\textwidth]{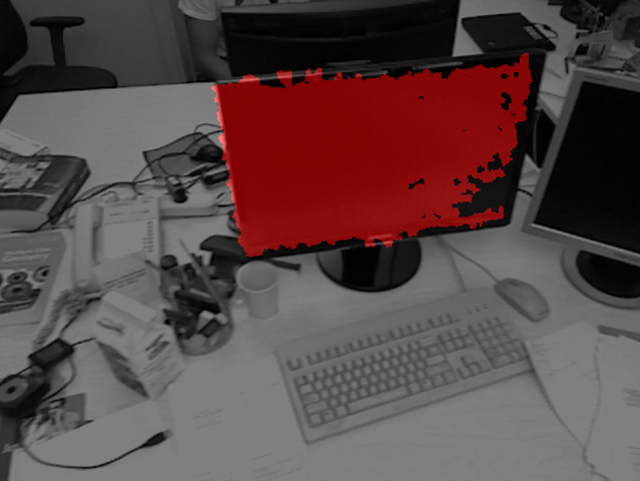}}~
\subfloat{\includegraphics[width=0.24\textwidth]{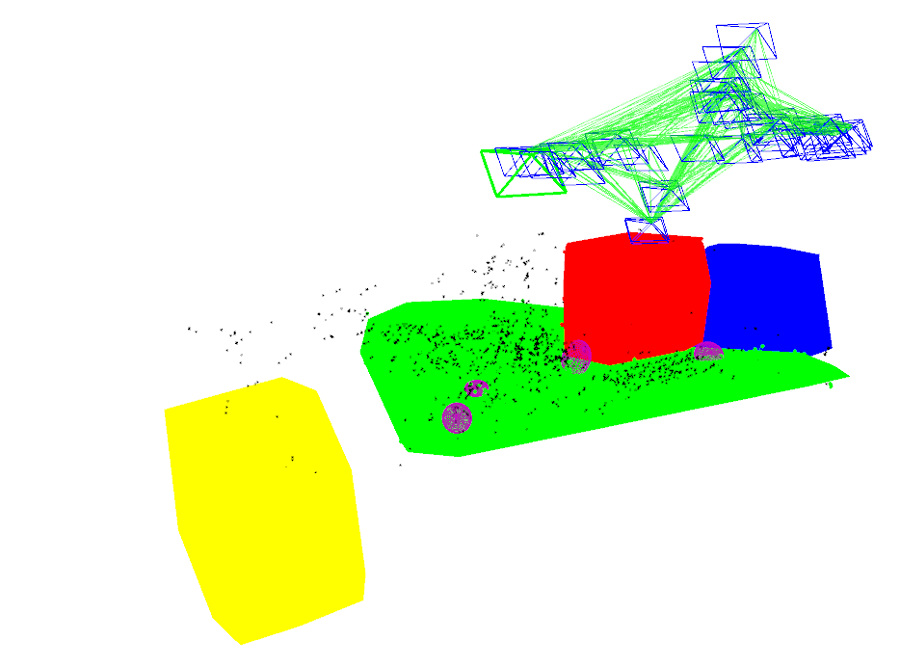}}~
\subfloat{\includegraphics[width=0.24\textwidth]{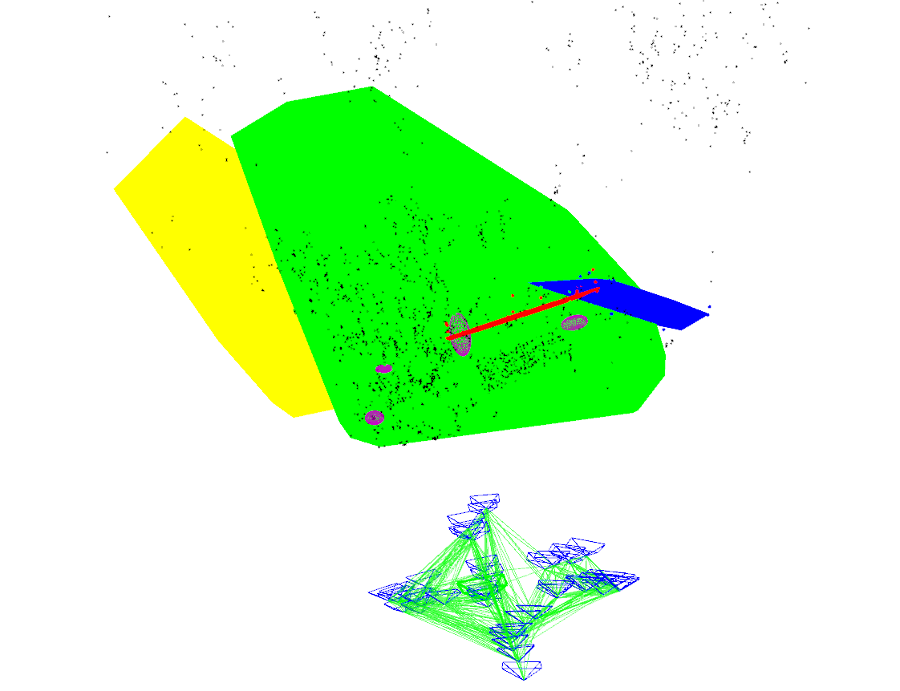}}\\[-2ex]
\subfloat{\includegraphics[width=0.24\textwidth]{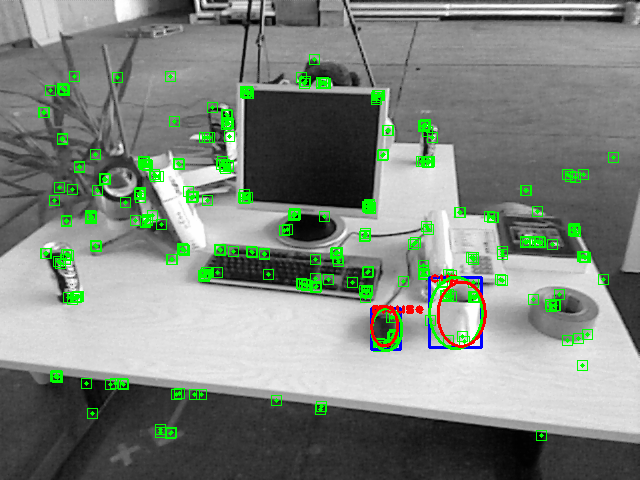}}~
\subfloat{\includegraphics[width=0.24\textwidth]{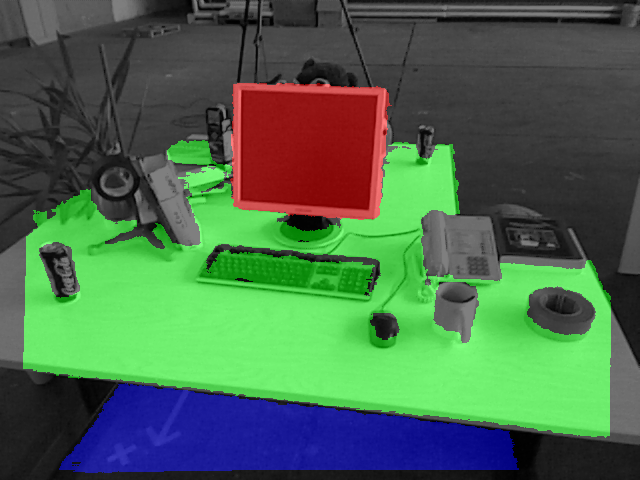}}~
\subfloat{\includegraphics[width=0.24\textwidth]{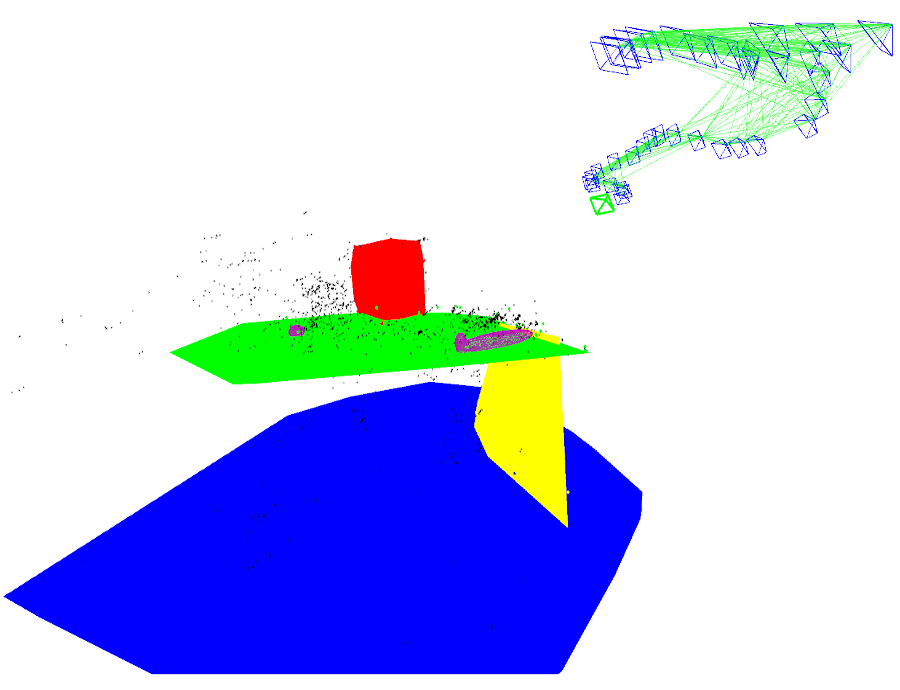}}~
\subfloat{\includegraphics[width=0.24\textwidth]{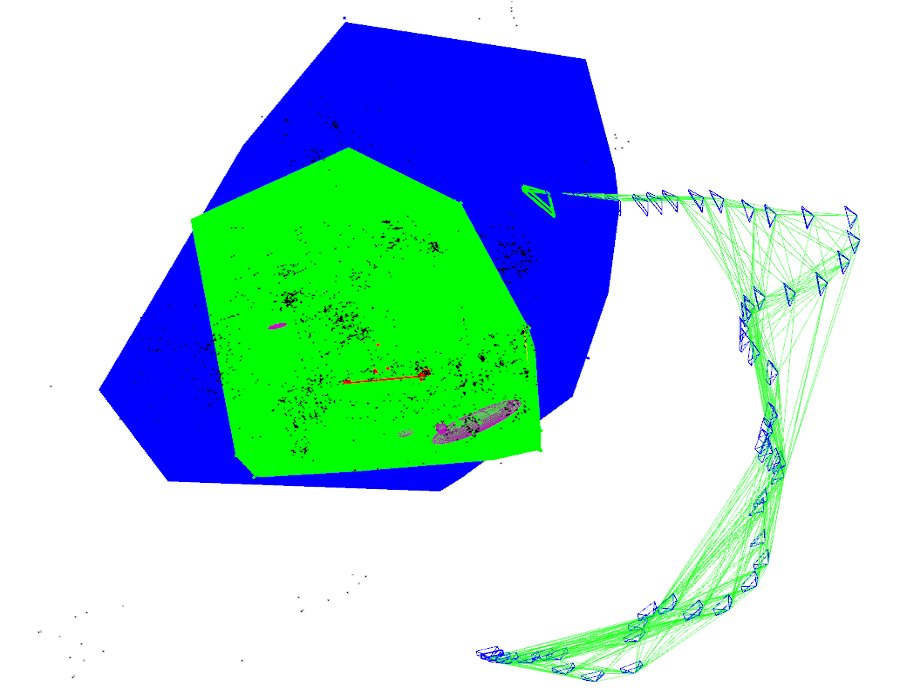}}\\[-2ex]
\subfloat{\includegraphics[width=0.24\textwidth]{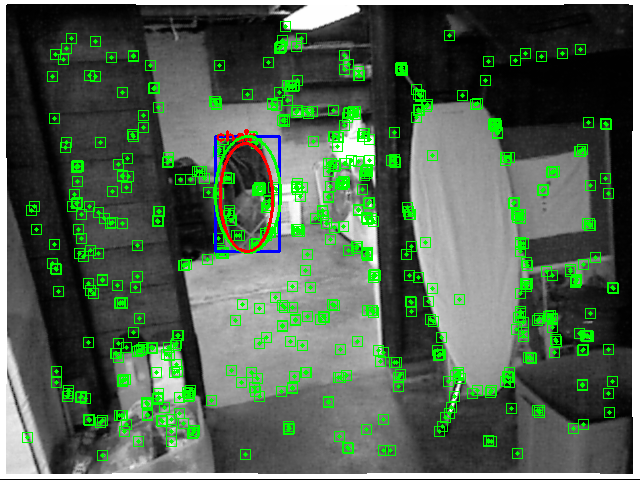}}~
\subfloat{\includegraphics[width=0.24\textwidth]{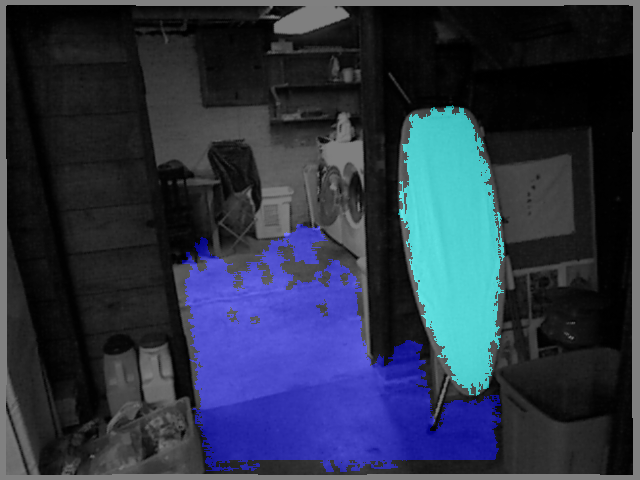}}~
\subfloat{\includegraphics[width=0.24\textwidth]{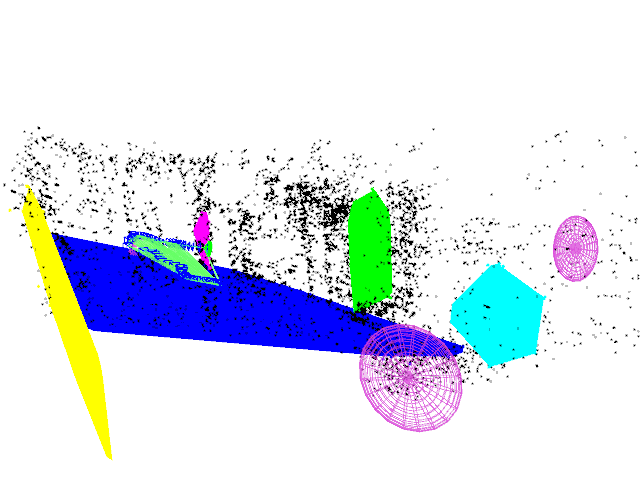}}~
\subfloat{\includegraphics[width=0.24\textwidth]{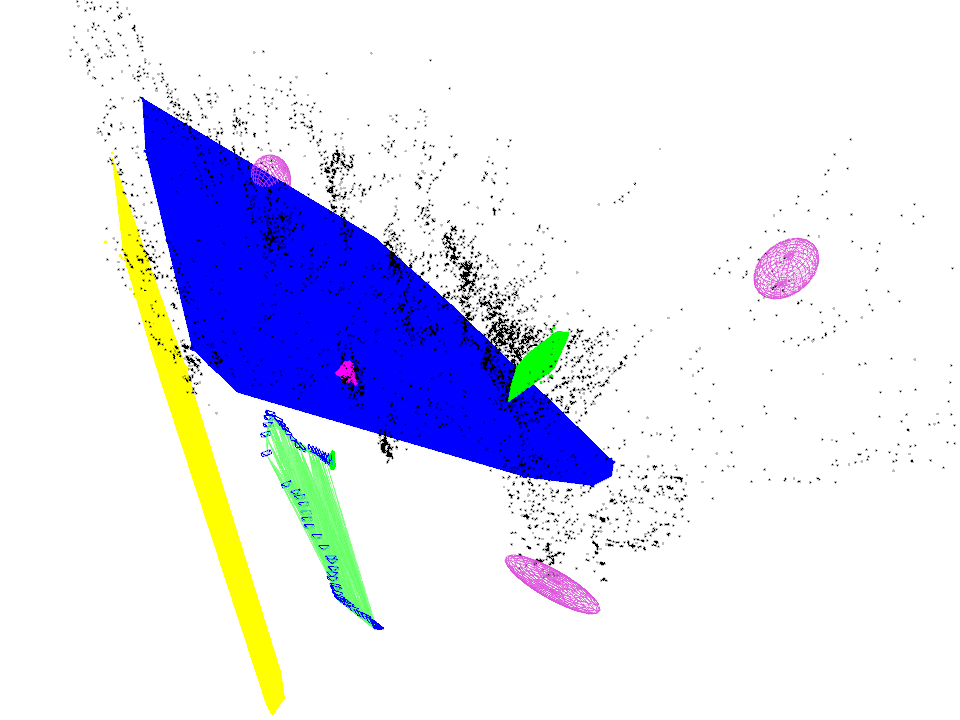}}\\[-2ex]
\setcounter{subfigure}{0}
\subfloat[Features \& Objects]{\includegraphics[width=0.24\textwidth]{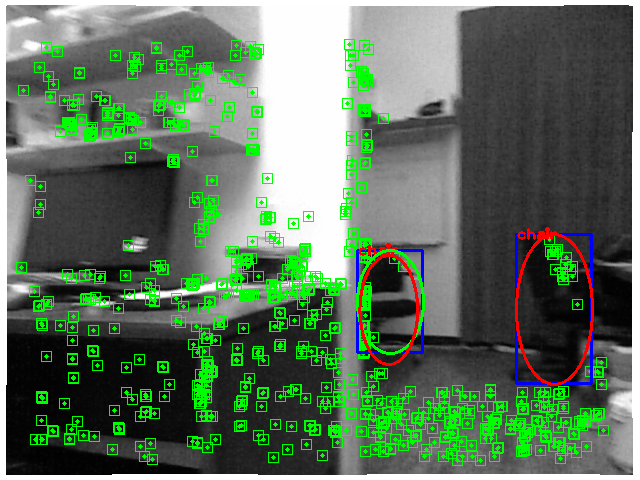}}~
\subfloat[Detected Planes]{\includegraphics[width=0.24\textwidth]{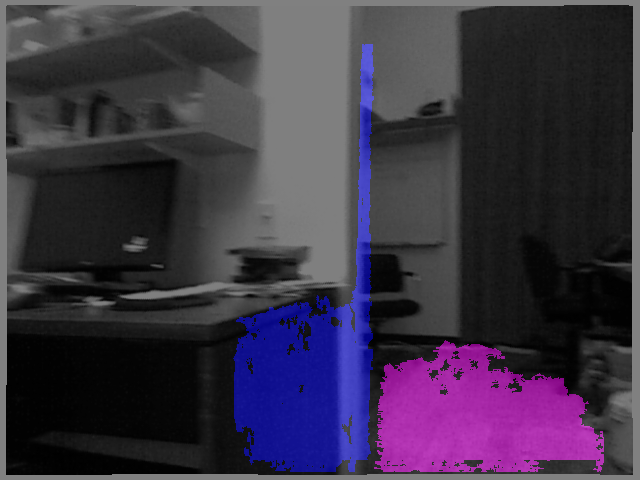}}~
\subfloat[Map (Side)]{\includegraphics[width=0.24\textwidth]{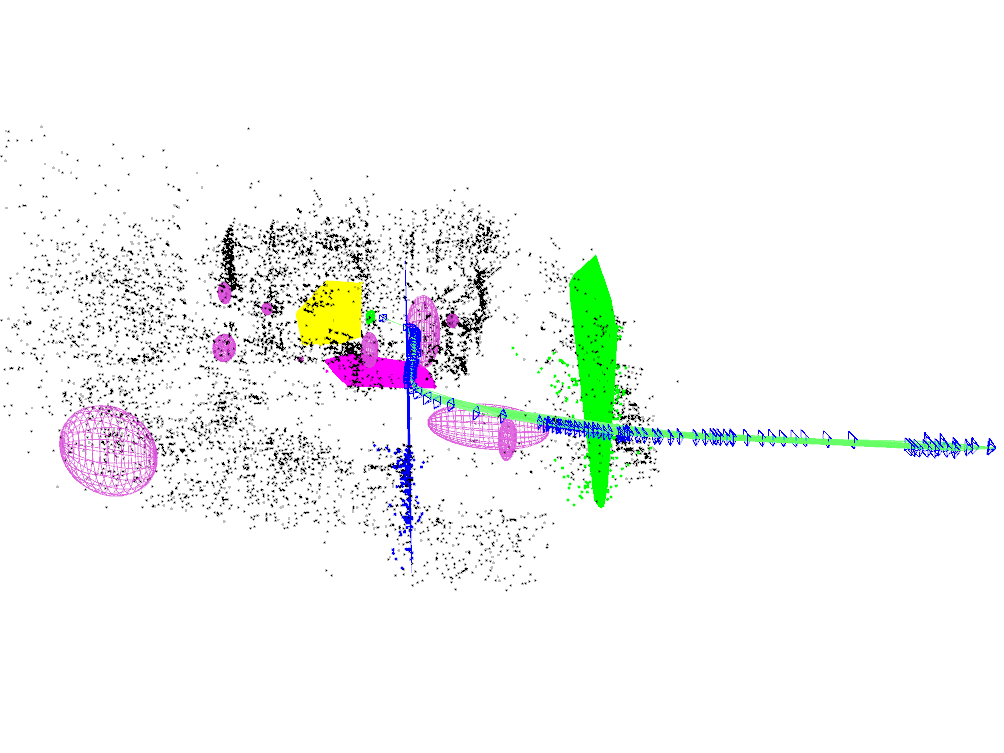}}~
\subfloat[Map (Top)]{\includegraphics[width=0.24\textwidth]{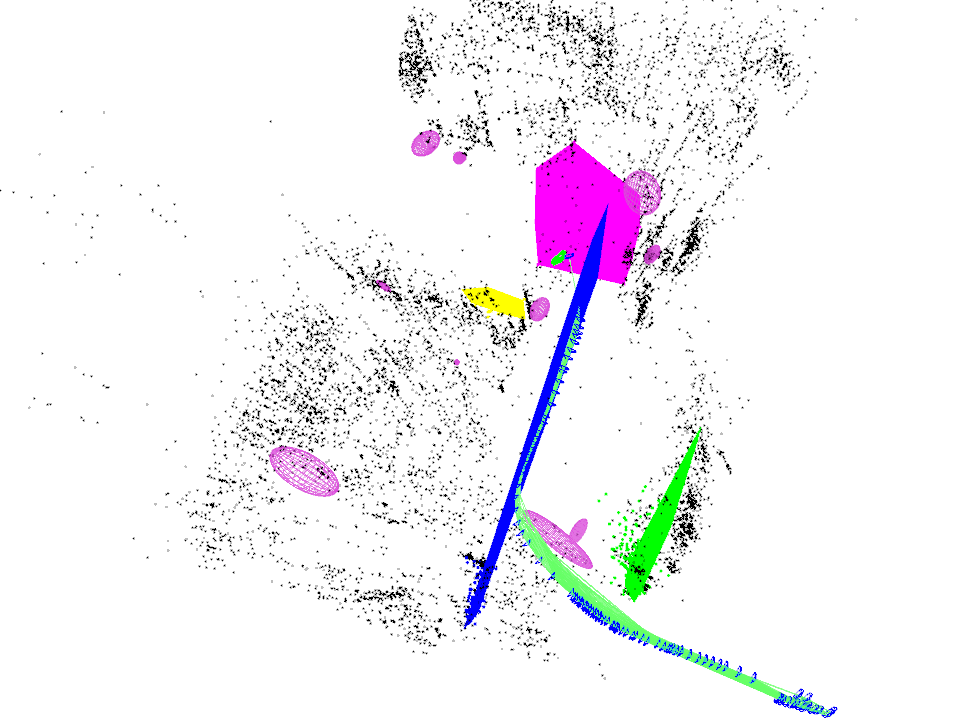}}
\caption{\small{Qualitative results for 2 different TUM RGBD and 2 different NYU-Depth-V2 datasets with rich planar structures and objects supported by planes}}
\label{fig:objects_experiments} 
\end{figure}

To show the quality of the mapping and tracking with planes and objects along with the Manhattan and supporting constraints, we use the sequences \texttt{fr1/xyz}, 
\texttt{fr2/desk} from TUM dataset, and \texttt{nyu/basement\_1a}, \texttt{nyu/office\_1} from NYU dataset.
The reconstructions are shown in Fig.~\ref{fig:objects_experiments}. 
The reconstructed map of \texttt{fr1/xyz} is depicted in column (c) and (d) of the first row. 
The planar structure of the map is consistent with the ground truth scene which consists of two planar monitors orthogonal to the green desk.
Quadrics corresponding to objects on the desk have been reconstructed tangent to the desk, their supporting plane. 
Column (a) shows tracked ORB features and detected COCO objects with confidence of at least 0.95 at the corresponding frame. The red ellipses in column (a) are the projection of the reconstructed quadric objects. They closely fit the detected blue bounding boxes and their corresponding green computed ellipses. 

\begin{figure}[t]
\centering
\setcounter{subfigure}{0}
\subfloat[\tiny{Generated Map (Before--Top)}]{\includegraphics[width=0.3\textwidth]{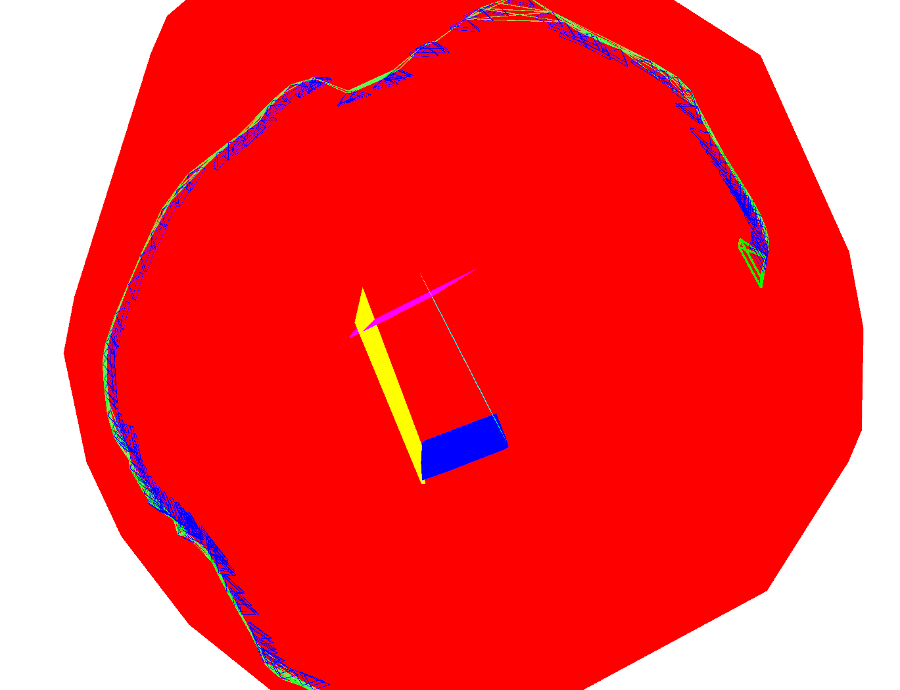}}~
\subfloat[\tiny{Generated Map (After--Top)}]{\includegraphics[width=0.3\textwidth]{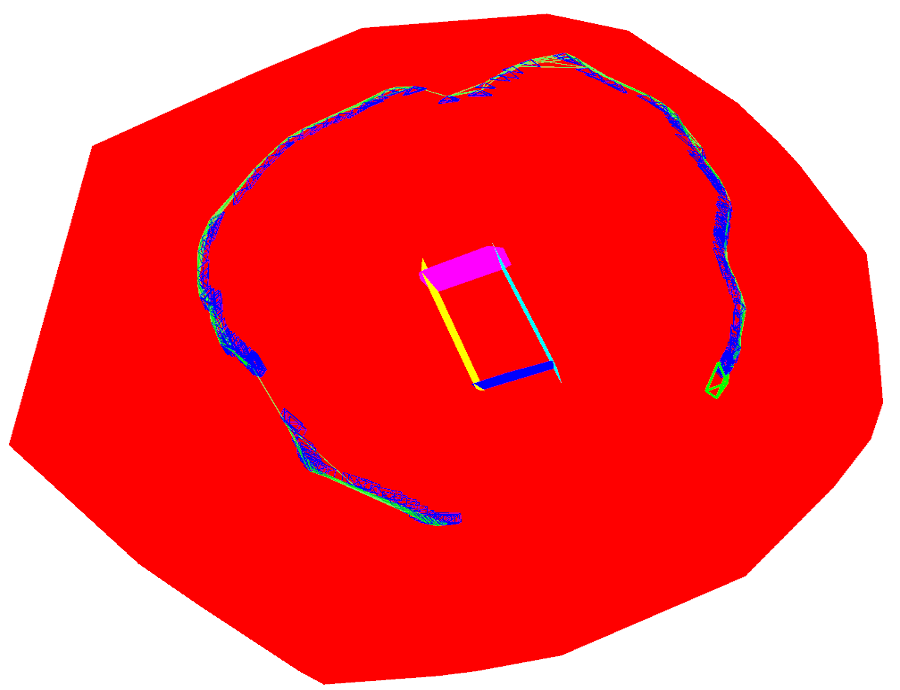}}~
\subfloat[\tiny{Generated Map (After--Side)}]{\includegraphics[width=0.3\textwidth]{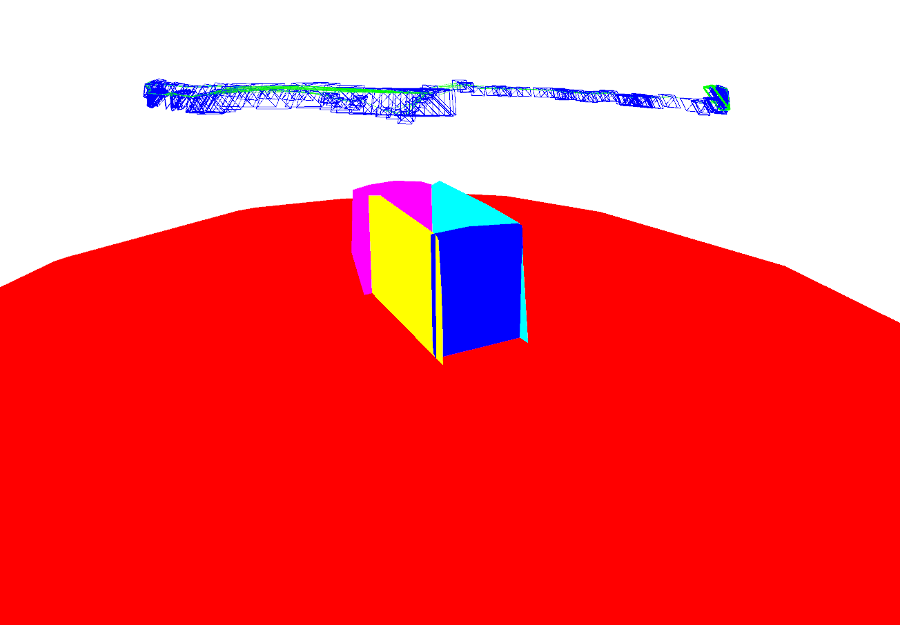}}
\caption{\small{Qualitative comparison of the reconstructed planes representing \texttt{cabinet} before and after imposing Manhattan assumption between the planes in the TUM \texttt{fr3/cabinet} dataset. Points and top-side plane of the cabinet have not been rendered for clarifying the difference in the map}}
\label{fig:manhattan_experiment} 
\end{figure}

We use \texttt{fr3/cabinet} to show the importance of using the Manhattan constraint.
The sequence contains a loop around a cabinet. All the faces of this cabinet are parallel or perpendicular to each other. Fig.~\ref{fig:manhattan_experiment} demonstrates the difference in the quality of the reconstruction of the cabinet's sides with and without Manhattan assumption in column (a) and column (b) respectively.

\begin{figure}[t]
\centering
\setcounter{subfigure}{0}
\subfloat[\tiny{Generated Map (Before)}]{\includegraphics[width=0.38\textwidth]{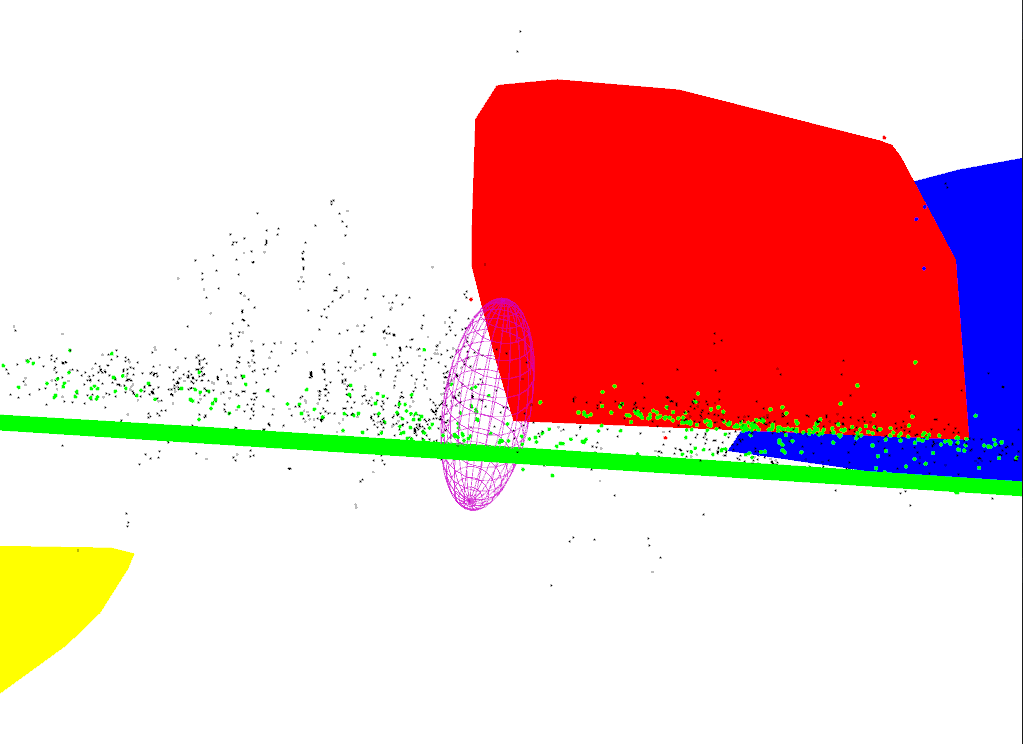}}~
\subfloat[\tiny{Generated Map (After)}]{\includegraphics[width=0.38\textwidth]{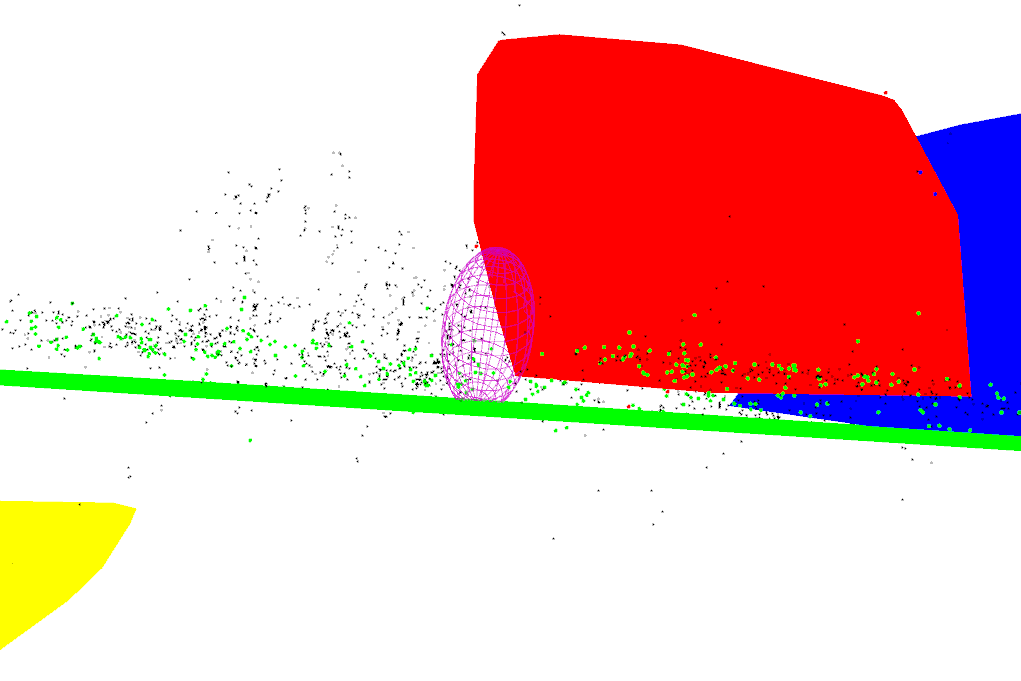}}
\caption{\small{Qualitative comparison of the reconstructed quadric representing object \texttt{cup} before and after imposing supporting/tangency constraint between the quadric and plane representing \texttt{desk}} in the TUM \texttt{fr1/xyz} dataset}
\label{fig:tangency_experiment} 
\end{figure}

Figs.~\ref{fig:tangency_experiment}(a,b) show the reconstructed quadric corresponding to the object on desk in the \texttt{fr1/xyz} before and after imposing the tangency constraint. Enforcing the tangency constraints makes sure that the quadric object is tangent to the supporting plane.

\subsection{Quantitative Comparison}
We compare the performance of the proposed SLAM system against the RGB-D variant of the state-of-the-art system ORB-SLAM2 for TUM RGBD dataset that the ground-truth trajectories are available. This baseline is a monocular point-based system that uses the depth information in the D-channel to initialize 3D points. Our implementation builds directly on their open-source C++ code-base, and we structure our results as an ablation study, considering the effects of introducing different landmarks and constraints. In each case, we report the RMSE Absolute Trajectory Error (ATE)\footnote{Comparison for RMSE of relative errors, RTE and RRE, as well as run-time analysis are reported in the supplementary material.} in Table \ref{tab:errors}.

We first consider the case where points are augmented by the plane information (\texttt{PP}).
This already improves the ATE in each case over the baseline,
which improves even further by enforcing Manhattan constraints (\texttt{PP+M}).
The Manhattan constraint significantly reduces the trajectory error when dominant structure is present in the scene.

Some sequences do not contain objects similar to the COCO dataset. For those that do,
we investigate using the combination of points and quadrics (\texttt{PQ}) as landmarks. While this reduces the trajectory drift compared to baseline, the improvement is smaller compared with using PP+M. 
Finally, we report numbers for the full system (\texttt{PPQ+MS}) in which points, planes and quadrics are used as landmarks and Manhattan and support constraints are enforced.
For \texttt{fr3/long\_office} the improvement is significant (51.07\%) because of the presence of a large loop in this sequence, where all of the points, planes and quadrics landmarks participate and are updated based on the loop closure. 

\begin{table}[t]
	\centering
    	\caption{
        \small{Comparison against RGB-D ORB-SLAM2. \texttt{PP}, \texttt{PP+M}, \texttt{PQ}, and \texttt{PPQ+MS} mean points-planes only, points-planes+Manhattan constraint, points-quadrics only, and all of the landmarks with Manhattan and supporting constraints, respectively. RMSE is reported for ATE in \texttt{cm} for 10 sequences in TUM RGBD datasets. Numbers in bold in each row represent the best performance for each sequence. Numbers in [~] show the percentage of improvement over ORB-SLAM2.}
        }
	\label{tab:errors}
	\begin{tabular*}{\textwidth}{l @{\extracolsep{\fill}} c|c|c|c|c}
	\hline
          Dataset            		& ORB-SLAM2&         PP   			&    	        PP+M					  &	      PQ 		  &		  PPQ+MS      
          \\\hline
	 \texttt{fr1/floor}      		& 1.4399   & 		1.3798          &  \textbf{1.3246} \scriptsize{[8.01\%]}  & 		 ---          &        ---   	   \\\hline 
	 \texttt{fr3/cabinet}    		& 7.9602   & 		7.3724          &  \textbf{2.1675} \scriptsize{[72.77\%]} & 		 ---		  &  	   ---   	   \\\hline 
	 \texttt{fr3/str\_notex\_near} 	& 1.6882   & 		1.0883  		&  \textbf{1.0648} \scriptsize{[36.93\%]} & 		 ---   		  &  	   ---   	   \\\hline 
	 \texttt{fr3/str\_notex\_far}  	& 2.0007   & 		1.9092     		&  \textbf{1.3722} \scriptsize{[31.41\%]} & 		 ---	      &  	   ---   	   \\\hline 
	 \texttt{fr1/xyz}        		& 1.0457   & 		0.9647          & 		0.9231  	& 	   	0.9544        &  \textbf{0.9038} \scriptsize{[13.57\%]}  \\\hline 
	 \texttt{fr1/desk}      		& 2.2668   & 		1.5267  		&  		1.4831  	& 		1.9821   	  &  \textbf{1.4029} \scriptsize{[38.11\%]}  \\\hline 
	 \texttt{fr2/xyz}      			& 0.3634   & 		0.3301          & 		0.3174    	& 		0.3453   	  &  \textbf{0.3097} \scriptsize{[14.78\%]}  \\\hline 
	 \texttt{fr2/rpy}      			& 0.3207   & 		0.3126          & 		0.3011    	& 		0.3195        &  \textbf{0.2870} \scriptsize{[10.51\%]}  \\\hline 
	 \texttt{fr2/desk}      		& 1.2962   & 		1.2031          & 		1.0186    	& 		1.1132        &  \textbf{0.8655} \scriptsize{[33.23\%]}  \\\hline 
	 \texttt{fr3/long\_office}    	& 1.5129   & 		1.0601 			&  		0.9902  	& 	   	1.3644   	  &  \textbf{0.7403} \scriptsize{[51.07\%]} \\\hline 
	\end{tabular*} 
\end{table}

\begin{figure}[hb]
\centering
\setcounter{subfigure}{0}
\subfloat[\tiny{\texttt{fr3/long\_office}}]{\includegraphics[width=0.35\textwidth]{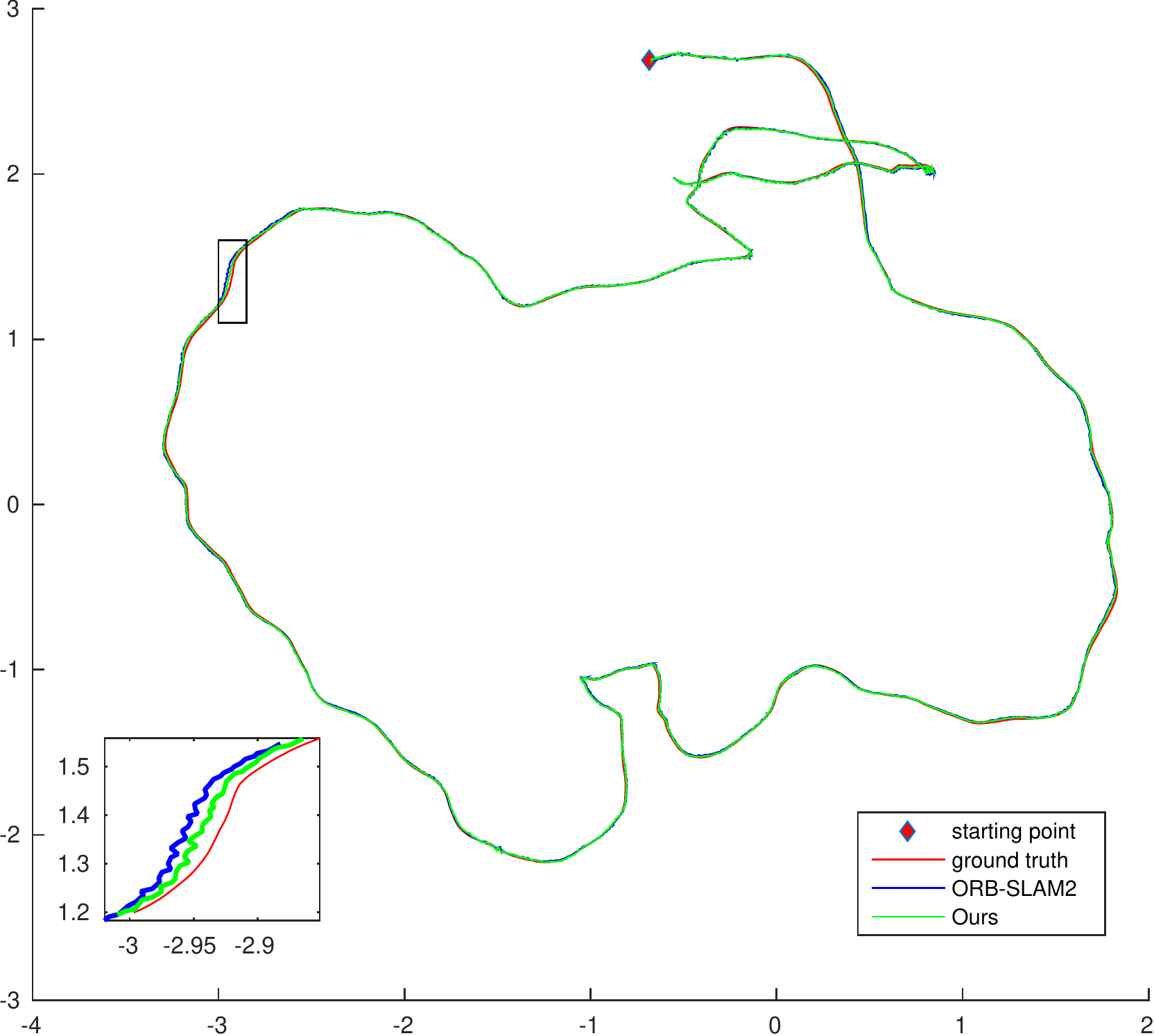}} \qquad \qquad
\subfloat[\tiny{\texttt{fr3/cabinet}}]{\includegraphics[width=0.35\textwidth]{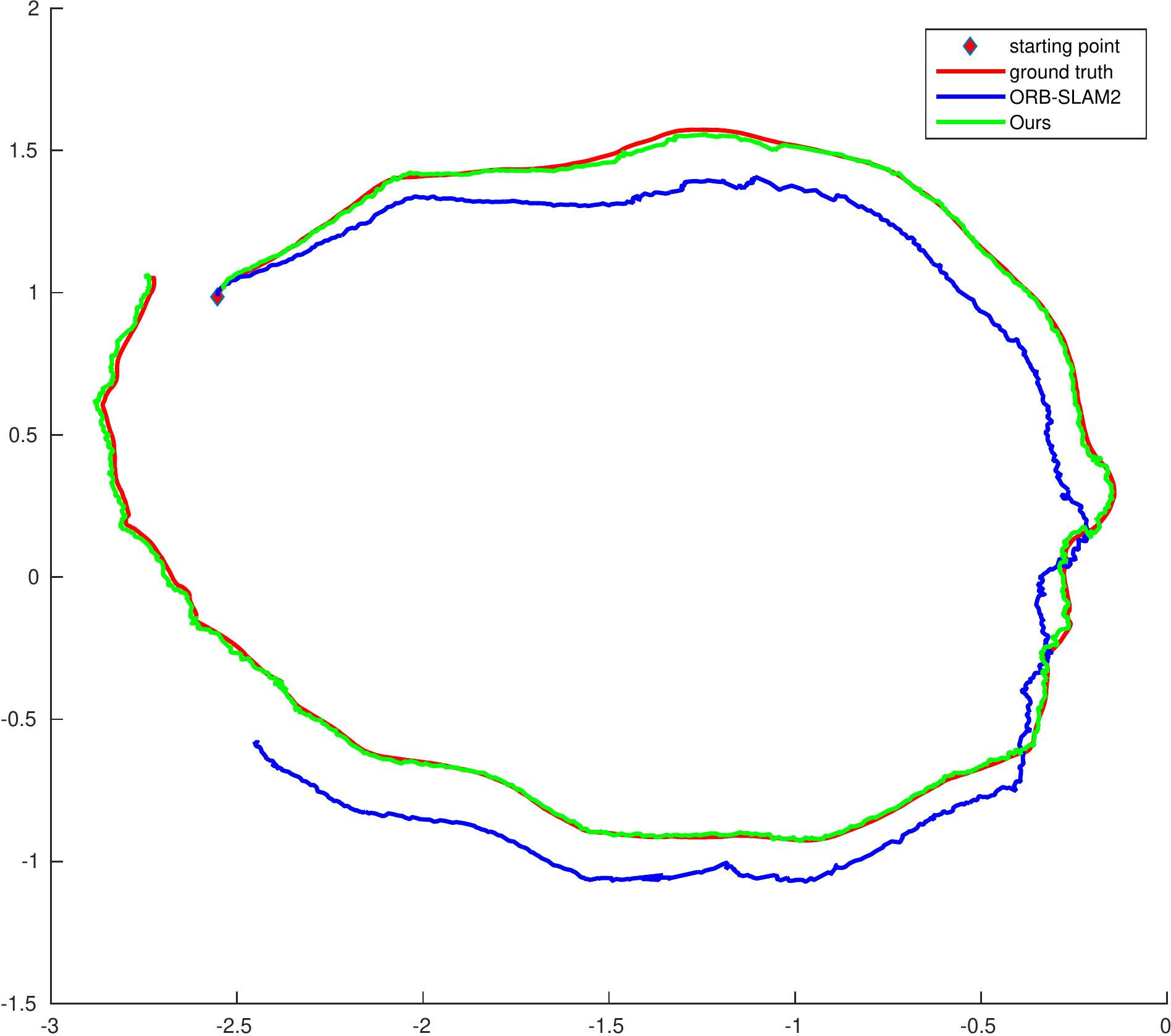}}
\caption{\small{Comparison of estimated trajectories of ORB-SLAM2, our system, and ground truth: \textbf{(a)} for TUM \texttt{fr3/long\_office} that has a large loop closure our trajectory is closer to the ground truth; \textbf{(b)} for TUM \texttt{fr3/cabinet} ORB-SLAM2 drifts significantly and loses track in this feature-poor sequence ($\sim$72\% improvement in ATE)}}
\label{fig:trajectories} 
\end{figure}

Comparison of the estimated trajectories of our system against ground truth is presented in Fig.~\ref{fig:trajectories} for two example TUM sequences. 

\section{Conclusions}\label{sec:conclusions}
In this work, we have explored the effects of incorporating planes and quadrics as higher-level geometric entities in a sparse point-based SLAM
framework. To do so we have introduced a new ellipsoid representation that is easily and effectively updated, and admits a simple method for imposing constraints between planes and objects. The improved performance due to using points and planes has been clearly shown by the experiments, most noticeably when there is dominant planar structure present. Of course in cases where enough planes are not present, the point based SLAM can still function as usual. 

Currently, the method works with RGB-D input. As in ``vanilla'' ORB-SLAM2, 3D map points are initialized with depth obtained from the D-channel of the RGB-D camera. We also use the D-channel to initialise planes, and this is both a bottleneck in terms of computation and presents a limitation on the sensor. In future, we will explore methods that can provide plane estimate from monocular input, which will enable us to transition to a purely monocular implementation. We also hope to further explore additional inter-object relations and introduce greater rigour to how and when the constraints are effected.

%
%
%
\bibliographystyle{splncs04}
\bibliography{refs}



\title{Supplementary of the paper: \\ Structure Aware SLAM using Quadrics and Planes\thanks{Supported by the ARC Laureate Fellowship FL130100102 to IR and the ACRV CE140100016.}}
\titlerunning{Supp: Structure Aware SLAM} 


\author{Mehdi Hosseinzadeh\inst{1,3} \and
Yasir Latif\inst{1,3} \and
Trung Pham\inst{4} \and
Niko Suenderhauf\inst{2,3} \and
Ian Reid\inst{1,3}}
%

\authorrunning{M. Hosseinzadeh et al.} 


\institute{The University of Adelaide, Adelaide, Australia 
\email{\{firstname.lastname\}@adelaide.edu.au}
\and 
Queensland University of Technology, Brisbane, Australia
\email{niko.suenderhauf@qut.edu.au}\\
\and
Australian Centre for Robotic Vision, Brisbane, Australia
\and
NVIDIA, Santa Clara CA 95051, USA \\
\email{trungp@nvidia.com}}

\maketitle


\section{Experiments}\label{sec:experiments}

In addition to the results in the main paper, the performance of our SLAM system is evaluated on more sequences from publicly available RGB-D TUM dataset, NYU-Depth V2 dataset, and also our own captured RGB-D sequence using the UR5 robot arm in our lab.

\subsection{Qualitative Results}

\subsubsection{TUM Dataset.}
\begin{figure}[ht]
\centering
\subfloat{\includegraphics[width=0.24\textwidth]{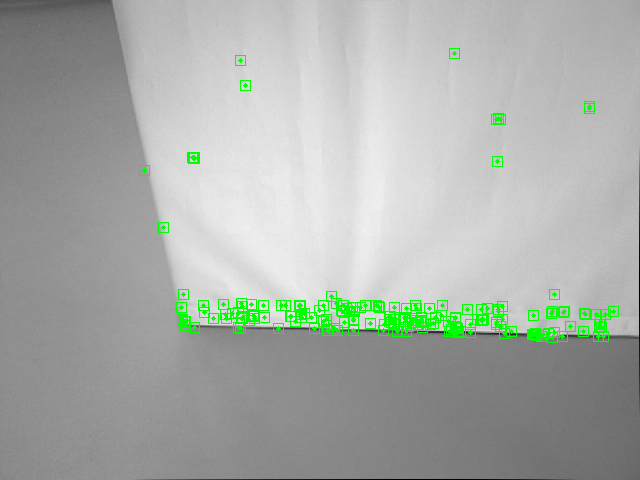}}~
\subfloat{\includegraphics[width=0.24\textwidth]{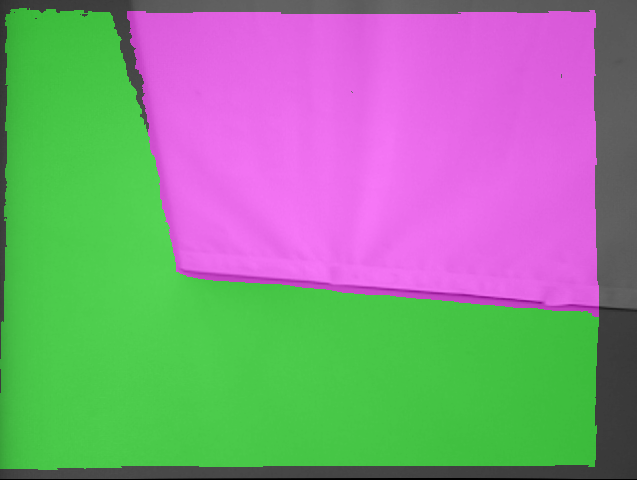}}~
\subfloat{\includegraphics[width=0.24\textwidth]{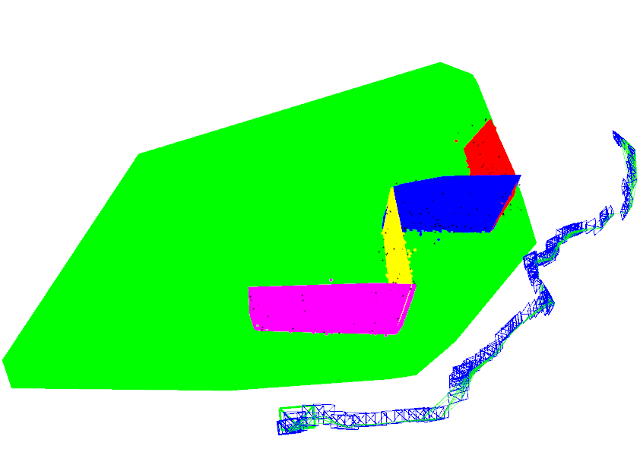}}~
\subfloat{\includegraphics[width=0.24\textwidth]{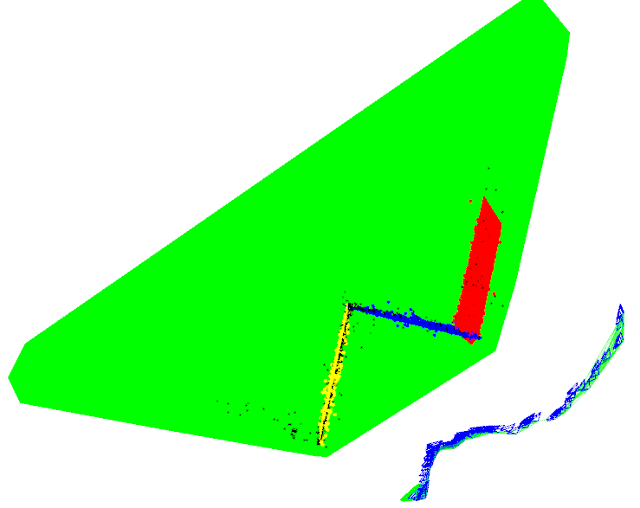}}\\[-2ex]
\setcounter{subfigure}{0}
\subfloat[Features \& Objects]{\includegraphics[width=0.24\textwidth]{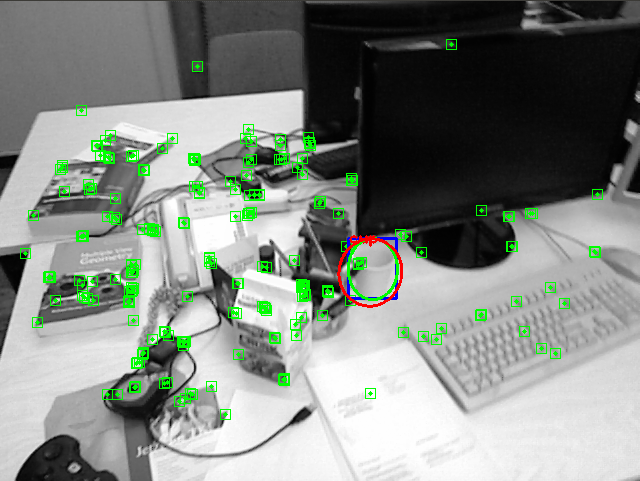}}~
\subfloat[Detected Planes]{\includegraphics[width=0.24\textwidth]{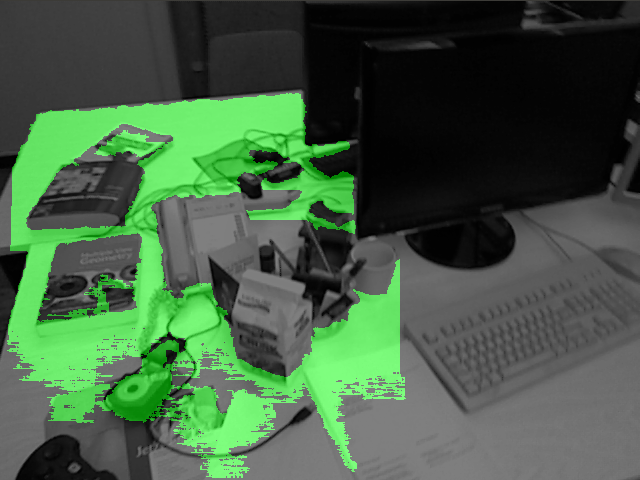}}~
\subfloat[Map (Side)]{\includegraphics[width=0.24\textwidth]{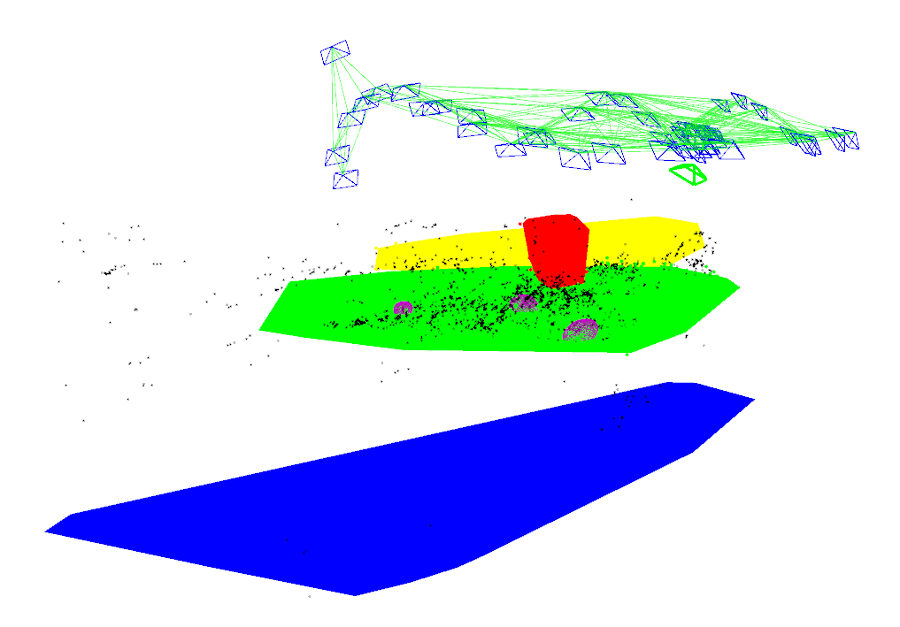}}~
\subfloat[Map (Top)]{\includegraphics[width=0.24\textwidth]{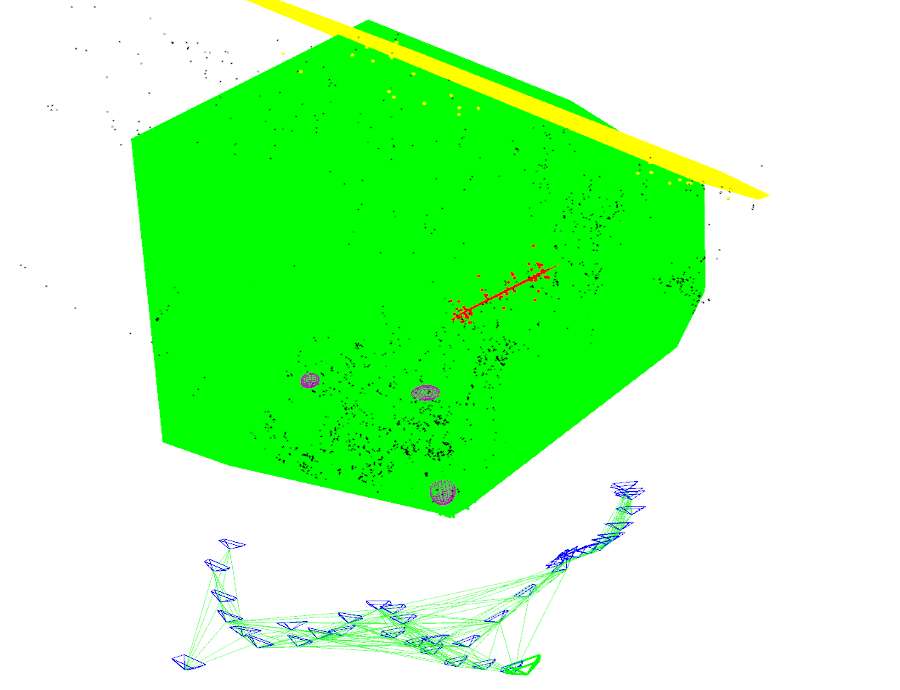}}
\caption{\small{Qualitative results for 2 different TUM RGB-D datasets with rich planar structures and also objects supported by planes}}
\label{fig:tum_experiments} 
\end{figure}

The results for extra sequences of \texttt{fr3/str\_notex\_near} and \texttt{fr1/desk} are illustrated in Fig.~\ref{fig:tum_experiments}. The figure shows the image frame along with tracked features and possible detected objects (column (a)), detected and segmented planes (column (b)), and the reconstructed map consisting of points, planes, and quadric objects from two different viewpoints (columns (c) and (d)). 

\subsubsection{UR5 Sequence.}
\begin{figure}[b]
\centering
\subfloat{\includegraphics[width=0.7\textwidth]{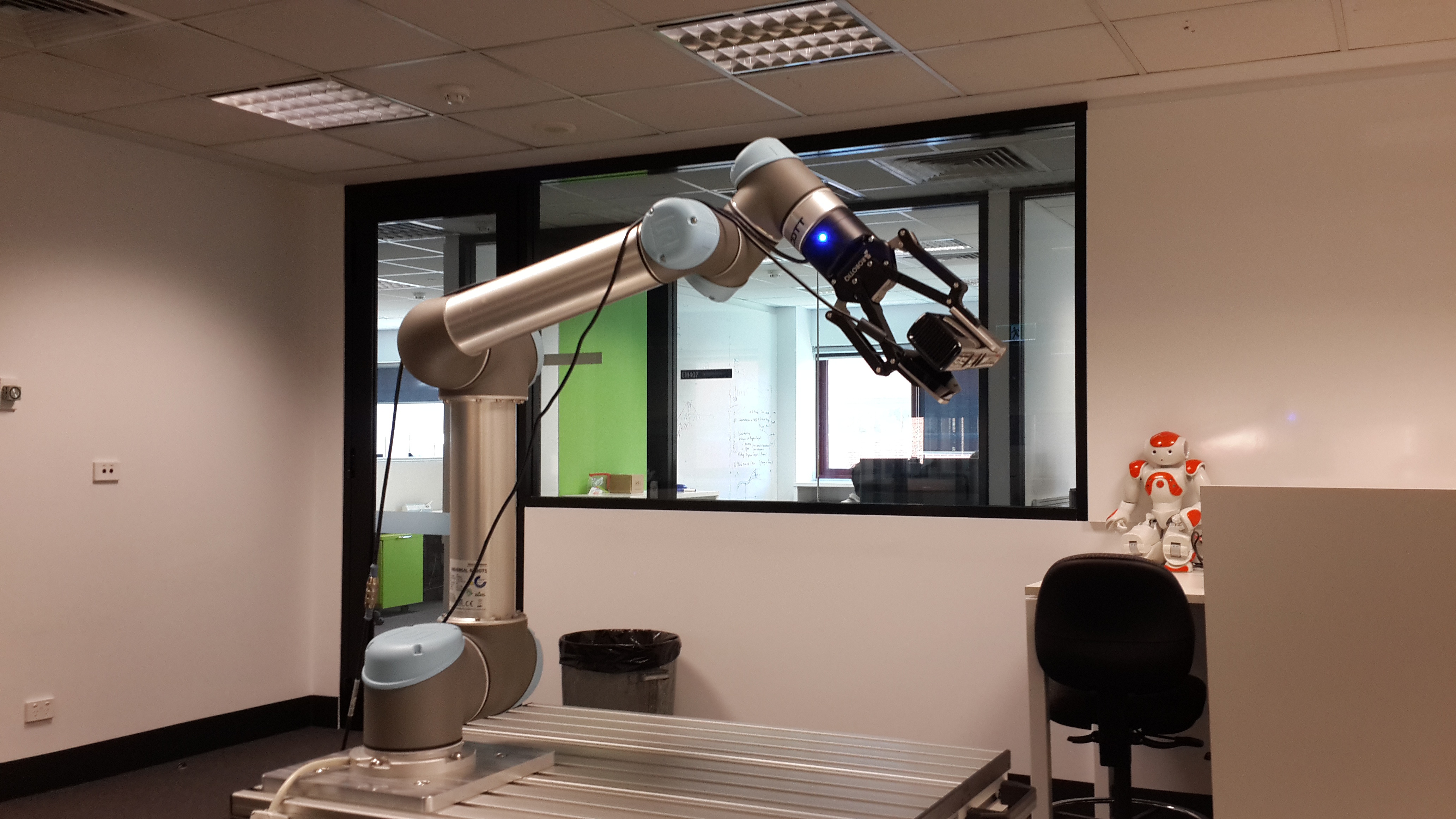}}
\caption{\small{The setup of the UR5 robot arm and Kinect in our lab, used to capture dataset}}
\label{fig:ur5_setup}
\end{figure}

To evaluate the reconstruction of our SLAM system on a sequence with more quadric objects, we captured an RGB-D sequence using the UR5 robot arm and Kinect. In this sequence, the camera is moved in a smooth trajectory over a table containing multiple objects.
The smooth motion of the robot allows us to avoid image blur and rolling shutter effects to achieve robust object detection.
The setup for UR5 robot arm is demonstrated in Fig.~\ref{fig:ur5_setup}. 

\begin{figure}[ht]
\centering
\subfloat{\includegraphics[width=0.3\textwidth]{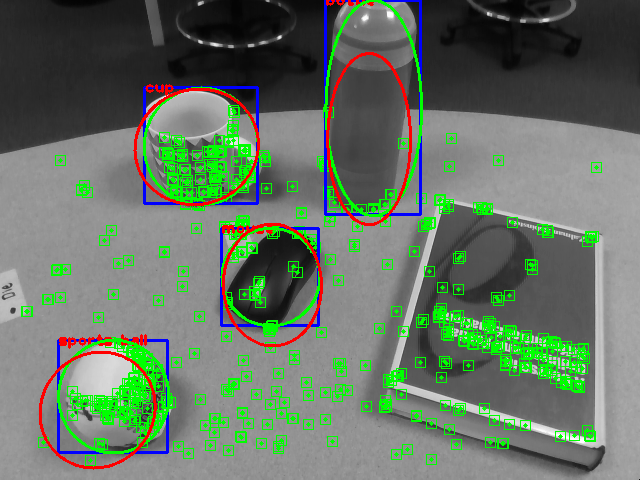}}~
\subfloat{\includegraphics[width=0.3\textwidth]{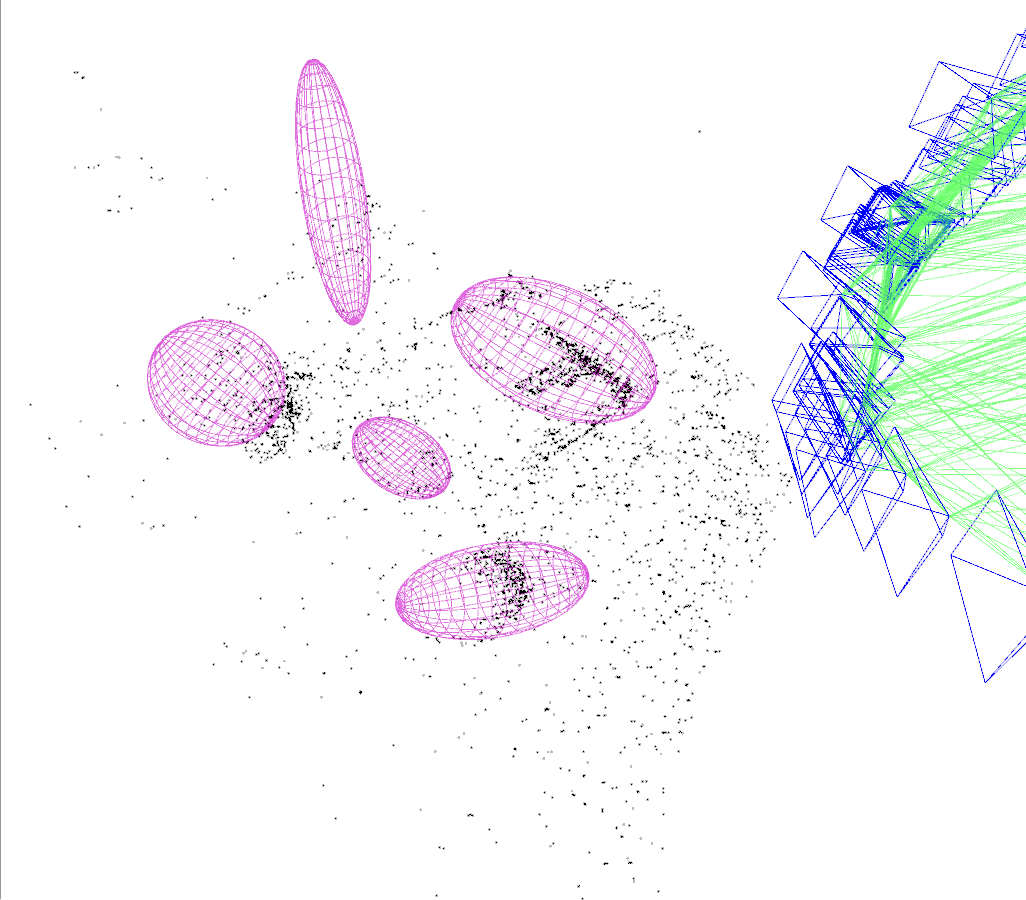}}\\[-4ex]
\setcounter{subfigure}{0}
\subfloat[Features \& Objects]{\includegraphics[width=0.3\textwidth]{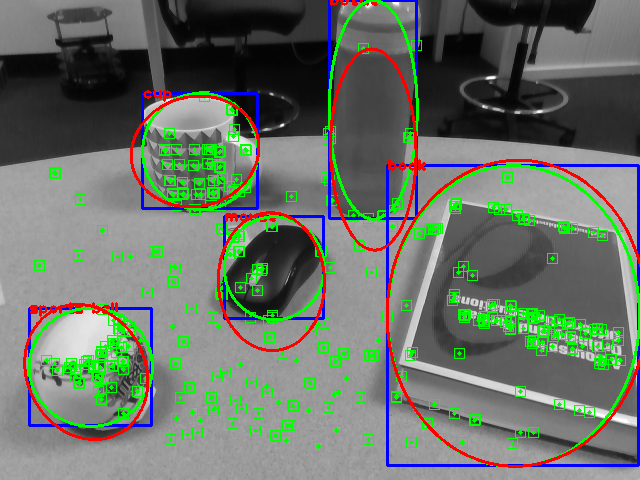}}~
\subfloat[Generated Map]{\includegraphics[width=0.3\textwidth]{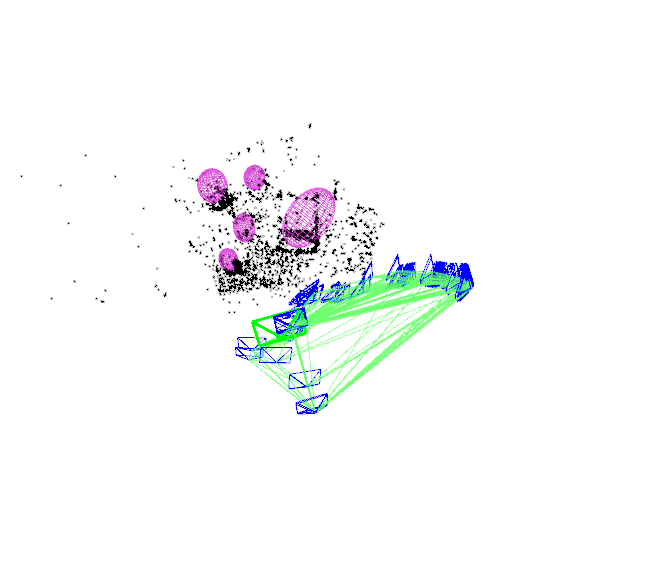}}
\caption{\small{Qualitative results for our captured dataset with UR5 robot}}
\label{fig:ur5_experiment} 
\end{figure}

The detected objects in two different frames of our UR5 sequence, as well as the reconstructed map are shown in Fig.~\ref{fig:ur5_experiment}. In this sequence, no planes are detected
therefore the map consists of points and quadric objects as landmarks without any additional constraints. 

\subsection{Quantitative Comparison}
The performance of our proposed SLAM system is compared against the RGB-D variant of the state-of-the-art system ORB-SLAM2, for TUM RGB-D dataset that the ground-truth trajectories are available. The results are structured as an ablation study, considering the effects of introducing different landmarks and constraints. Comparison for RMSE of Absolute Trajectory Error (ATE) is reported in the main paper and the comparisons for RMSE of Relative Translational Error (RTE), and Relative Rotational Error (RRE) are reported in Table~\ref{tab:errors_rte}, and Table~\ref{tab:errors_rre}, respectively.

\begin{table}
	\centering
   	\caption{
       \small{Comparison against RGB-D ORB-SLAM2. \texttt{PP}, \texttt{PP+M}, \texttt{PQ}, and \texttt{PPQ+MS} signify points-planes only, points-planes+Manhattan constraint, points-quadrics only, and all of the landmarks with Manhattan and supporting constraints, respectively. RMSE is reported for RTE in \texttt{cm} for 10 sequences in TUM RGBD datasets. Numbers in bold in each row represent the best performance for each sequence. Numbers in [~] show the percentage of improvement over ORB-SLAM2.} 
       }
	\label{tab:errors_rte}
	\begin{tabular*}{\textwidth}{l @{\extracolsep{\fill}} c|c|c|c|c}
	\hline
         Dataset            		& ORB-SLAM2&         PP   			&    	        PP+M					  &	      PQ 		  &		  PPQ+MS      
         \\\hline
	 \texttt{fr1/floor}      		& 4.2161   & 		3.8789          &  \textbf{3.6381} \scriptsize{[13.71\%]}  & 		 ---          &        ---   	   \\\hline 
	 \texttt{fr3/cabinet}    		& 15.1002  & 		14.4081         &  \textbf{5.1328} \scriptsize{[66.01\%]} & 		 ---		  &  	   ---   	   \\\hline 
	 \texttt{fr3/str\_notex\_near} 	& 4.0383   & 		2.4540  		&  \textbf{2.3533} \scriptsize{[41.73\%]} & 		 ---   		  &  	   ---   	   \\\hline 
	 \texttt{fr3/str\_notex\_far}  	& 3.4869   & 		3.3523     		&  \textbf{3.0834} \scriptsize{[11.57\%]} & 		 ---	      &  	   ---   	   \\\hline 
	 \texttt{fr1/xyz}        		& 1.5693   & 		1.4675          & 		1.2876  	& 	   	1.3795        &  \textbf{1.2464} \scriptsize{[20.38\%]}  \\\hline 
	 \texttt{fr1/desk}      		& 4.0835   & 		3.2994  		&  		3.1174  	& 		3.7453   	  &  \textbf{3.0237} \scriptsize{[25.95\%]}  \\\hline 
	 \texttt{fr2/xyz}      			& 1.2107   & 		1.0765          & 		0.9659    	& 		1.1964   	  &  \textbf{0.9309} \scriptsize{[23.11\%]}  \\\hline 
	 \texttt{fr2/rpy}      			& 0.5534   & 		0.5322          & 		0.5073    	& 		0.5484        &  \textbf{0.4883} \scriptsize{[11.76\%]}  \\\hline 
	 \texttt{fr2/desk}      		& 4.7783   & 		4.7110          & 		3.6209    	& 		4.6309        &  \textbf{3.5545} \scriptsize{[25.61\%]}  \\\hline 
	 \texttt{fr3/long\_office}    	& 3.0555   & 		2.6223 			&  		2.4750  	& 	   	2.5887   	  &  \textbf{1.8906} \scriptsize{[38.12\%]} \\\hline 
	\end{tabular*} 
\end{table}

\begin{table}
	\centering
   	\caption{
       \small{Comparison against RGB-D ORB-SLAM2. \texttt{PP}, \texttt{PP+M}, \texttt{PQ}, and \texttt{PPQ+MS} signify points-planes only, points-planes+Manhattan constraint, points-quadrics only, and all of the landmarks with Manhattan and supporting constraints, respectively. RMSE is reported for RRE in \texttt{deg} for 10 sequences in TUM RGBD datasets. Numbers in bold in each row represent the best performance for each sequence. Numbers in [~] show the percentage of improvement over ORB-SLAM2.} 
       }
	\label{tab:errors_rre}
	\begin{tabular*}{\textwidth}{l @{\extracolsep{\fill}} c|c|c|c|c}
	\hline
         Dataset            		& ORB-SLAM2&         PP   			&    	        PP+M					  &	      PQ 		  &		  PPQ+MS      
         \\\hline
	 \texttt{fr1/floor}      		& 3.3229   & 		2.8856          &  \textbf{2.7839} \scriptsize{[16.22\%]}  & 		 ---          &        ---   	   \\\hline 
	 \texttt{fr3/cabinet}    		& 6.8639   & 		6.5623          &  \textbf{2.9125} \scriptsize{[57.57\%]} & 		 ---		  &  	   ---   	   \\\hline 
	 \texttt{fr3/str\_notex\_near} 	& 1.8476   & 		1.1541  		&  \textbf{1.1125} \scriptsize{[39.79\%]} & 		 ---   		  &  	   ---   	   \\\hline 
	 \texttt{fr3/str\_notex\_far}  	& 0.8479   & 		0.7679     		&  \textbf{0.6695} \scriptsize{[21.04\%]} & 		 ---	      &  	   ---   	   \\\hline 
	 \texttt{fr1/xyz}        		& 0.9871   & 		0.9534          & 		0.9037  	& 	   	0.9433        &  \textbf{0.8822} \scriptsize{[10.63\%]}  \\\hline 
	 \texttt{fr1/desk}      		& 1.8547   & 		1.7817  		&  		1.6932  	& 		1.8044   	  &  \textbf{1.5214} \scriptsize{[17.97\%]}  \\\hline 
	 \texttt{fr2/xyz}      			& 0.5036   & 		0.4888          & 		0.4456    	& 		0.4904   	  &  \textbf{0.4308} \scriptsize{[14.46\%]}  \\\hline 
	 \texttt{fr2/rpy}      			& 0.9667   & 		0.9586	        & 		0.9452    	& 		0.9611        &  \textbf{0.8770} \scriptsize{[9.28\%]}  \\\hline 
	 \texttt{fr2/desk}      		& 1.6062   & 		1.4232          & 		1.1236    	& 		1.2559        &  \textbf{1.0034} \scriptsize{[37.53\%]}  \\\hline 
	 \texttt{fr3/long\_office}    	& 0.8927   & 		0.8062 			&  		0.7109  	& 	   	0.8744   	  &  \textbf{0.6230} \scriptsize{[30.21\%]} \\\hline 
	\end{tabular*} 
\end{table}

\section{Runtime Analysis}

All the experiments of our SLAM system have been carried out by a commodity machine with an Intel Core i7-4790 CPU at 3.6 GHz. All the source code is implemented in C$++$. 

In terms of runtime, the bottle-neck of the system is the object detection component that is based on Faster-RCNN which operates at less than 10 frames-per-second (more than 100 msec per frame). 
Therefore, the object detections have been pre-evaluated for all of the sequences and the results of the per-frame object detection have been fed to the system during online operation.
This is not a fundamental restriction of our system and in future will be alleviated by incorporating a real-time object detection method.

The runtime analysis and average statistics of different components and threads of our SLAM system evaluated on RGB-D TUM and NYU-Depth-V2 datasets are shown in Table~\ref{tab:runtime}. 
The system consists of three parallel modules: tracking, local map updates, and global map update when a loop is closed. 
The tracking thread has to run at frame-rate while the other two can operate at a slower pace.
Plane segmentation is done per-frame to do data-association against planes present in the map.
The reported numbers are for the full system that utilizes all the landmarks (points, planes, quadric objects). The local map optimisation is carried out in a parallel thread after creating and adding a keyframe to the map. 

\begin{table}[t]
    \centering
    \caption{\small{Average runtime statistics of different components and threads of our SLAM system evaluated on the RGB-D TUM and NYU-Depth-V2 datasets with all of the landmarks (points, planes, and quadric objects) with Manhattan assumption and supporting constraints.}} 
    \label{tab:runtime}
    \bigskip
    \begin{tabular*}{0.6\textwidth}{l|c}
    \hline
	  Main Components and Threads 				&   Runtime (msec)	 \\
            \hline
	\texttt{Plane Segmentation}					& 	  23.6 			 \\
	\texttt{Tracking \& Matching Landmarks}		& 	  27.1 			 \\
	\texttt{Local Mapping Optimization}			& 	  348.4 		 \\
	\texttt{Global Bundle Adjustment}			& 	  2170.6 		 \\
    \texttt{Average Frame Time}					& 	  51.9 			 \\
    \hline
    \end{tabular*}
\end{table}

\end{document}